\documentclass[11pt]{article}
\usepackage[colorlinks=true,pdfborder=001,linkcolor=blue,citecolor=blue]{hyperref}
\usepackage{interval}
\usepackage{url}
\usepackage{bbm}
\usepackage{stackengine}
\usepackage{nccmath}
\usepackage{mathtools}

\usepackage{mathrsfs}
\usepackage[title]{appendix}
\usepackage{adjustbox}
\usepackage{subcaption}
\usepackage{pgfplots}
\usepackage{tikz}
\usepackage{tikz,fullpage}
\usepackage{rotating}
\usepackage{tabu}
\usepackage{hhline}
\usepackage{multirow}
\usepackage{float}
\restylefloat{table}
\usepackage{adjustbox}
\usepackage[round]{natbib}
\usepackage{booktabs}
\usepackage{array}
\usepackage{tabularx}
\usepackage[utf8]{inputenc}
\usepackage{amssymb,amsmath}
\usepackage{amsfonts}
\usepackage{latexsym}
\usepackage[english]{babel}
\usepackage[T1]{fontenc}
\usepackage{eepic}
\usepackage{epic}
\usepackage{epsfig}
\usepackage{graphics,color}
\usepackage{verbatim}
\usepackage{lscape}
\usepackage{amsthm}
\usepackage{amsmath}
\usepackage{placeins}
\usepackage{float}
\usepackage{caption}
\usepackage{color}
\usepackage{setspace}
\usepackage[ruled,linesnumbered]{algorithm2e}
\usepackage{frcursive}

\usepackage{mathrsfs}
\usepackage{soul}
\usepackage{tikz}
\usepackage{tikz,fullpage}
\usetikzlibrary{arrows,petri,topaths,automata}

\usepackage{pgfplots}
\pgfplotsset{compat=1.18, set layers}
\usetikzlibrary{patterns}

\definecolor{myblue}{RGB}{29,114,184}
\definecolor{myorange}{RGB}{245,124,0}
\definecolor{lightgray}{RGB}{220,220,220}
\definecolor{mygreen}{RGB}{46,139,87}
\definecolor{mykhaki}{RGB}{225, 224, 175}

\definecolor{myblue}{RGB}{49, 130, 189}
\definecolor{myorange}{RGB}{253, 141, 60}
\definecolor{mylightgreen}{RGB}{161, 217, 155}
\definecolor{mypurple}{RGB}{158, 154, 200}
\definecolor{mytan}{RGB}{222, 189, 175}
\definecolor{mygray}{RGB}{115, 115, 115}
\definecolor{myyellow}{RGB}{237, 248, 177}
\definecolor{mycyan}{RGB}{179, 238, 244}

\makeatletter

\renewcommand*\thesection{\arabic{section}}

\newtheorem{exam}{Example}

\newif\ifbold
\boldfalse
\boldtrue
\newcommand{\bbf}{\ifbold\bgroup\bf\fi}
\newcommand{\ebf}{\ifbold\egroup\fi}

\renewcommand{\textbf}[1]{\begingroup\bfseries\mathversion{bold}#1\endgroup}
\makeatletter    
\renewcommand{\section}{\@startsection {section}{1}{\z@}%
             {-2ex \@plus -1ex \@minus -.2ex}%
             {1ex \@plus.2ex}%
             {\normalfont\Large\rmfamily\bfseries}}
\renewcommand{\subsection}{\@startsection{subsection}{2}{\z@}%
             {-1.25ex\@plus -1ex \@minus -.2ex}%
             {.75ex \@plus .2ex}%
             {\normalfont\large\rmfamily\bfseries}}
\def\@listI{\leftmargin\leftmargini       
            \parsep .25ex \@plus .1ex     
            \topsep .25ex \@plus .1ex     
            \itemsep \parsep}
\let\@listi\@listI
\@listi
\makeatother
\makeatletter

\@addtoreset{equation}{section}
\makeatother
\makeatletter

\@addtoreset{table}{section}
\makeatother

\usepackage[margin=1in]{geometry}
\usepackage{graphicx}
\graphicspath{ {./Figures/} }
\definecolor{purple}{rgb}{0.4,0.2,1}
\usepackage{titling}

\title{
  \LARGE\bf A Fair OR–ML Framework for Resource Substitution \\
  in Large-Scale Networks
\vspace{1ex}
}

\author{\large Ved Mohan,$^{1,*}$ El Mehdi Er Raqabi,$^{1}$ Pascal Van Hentenryck$^{1}$ \\
\footnotesize$^1$\emph{H. Milton Stewart School of Industrial and Systems Engineering, Georgia Institute of Technology, Atlanta, USA}\\
\footnotesize$^*$\emph{Corresponding Author: vedmohan@gatech.edu}\\
}

\date{}

\begin{document}
\maketitle

\vspace{2cm}
\begin{abstract}
\vspace{0.5cm}

Ensuring that the right resource is available at the right location and time remains a major challenge for organizations operating large-scale logistics networks. The challenge comes from uneven demand patterns and the resulting asymmetric flow of resources across the arcs, which create persistent imbalances at the network nodes. Resource substitution among multiple, potentially composite and interchangeable, resource types is a cost-effective way to mitigate these imbalances. This leads to the resource substitution problem, a complex optimization problem both in theory and practice, which aims at determining the minimum number of resource substitutions from an initial assignment to minimize the overall network imbalance. In decentralized settings, where distributed schedulers pursue local objectives, achieving globally coordinated solutions becomes even more difficult. When substitution entails costs, effective prescriptions must also incorporate fairness and account for the individual preferences of schedulers.

This paper presents a generic framework that combines operations research (OR) and machine learning (ML) to enable \emph{fair} resource substitution in large networks. The OR component models and solves the resource substitution problem under a fairness lens. The ML component leverages historical data to learn schedulers’ preferences, guide intelligent exploration of the decision space, and enhance computational efficiency by dynamically selecting the top-$\kappa$ resources for each arc in the network. The framework produces a portfolio of high-quality solutions from which schedulers can select satisfactory trade-offs. The proposed framework is applied to the network of one of the largest package delivery companies in the world, which serves as the primary motivation for this research. Computational results demonstrate substantial improvements over state-of-the-art methods, including an 80\% reduction in model size and a 90\% decrease in execution time while preserving optimality.

\vspace{0.2cm}
\par
\vspace{0.2cm}

\end{abstract}    

\newpage
\setlength{\parindent}{1em}
\setlength{\parskip}{0.5em}
\renewcommand{\baselinestretch}{1.5}
\doublespacing

\vspace{-3cm}

\section{Introduction}\label{section:1}

A major challenge faced by many organizations operating in large networks is to ensure that the right resource is available at the right location and at the right time. Achieving such allocation is complex partly because the flow of products within the network creates an imbalance. Any mileage between nodes that does not generate revenue, often called \emph{deadhead} or \emph{empty} miles, inflates costs and reduces productivity. Deadhead mileage across the American trucking industry was 16.3 percent of total mileage in 2023 \citep{intro_businessreport_atri}. Decision makers are forced to make empty moves to reposition resources. Heavy-duty trucks, including long-haul vehicles, account for approximately 25\% of carbon dioxide emissions from U.S. transportation, despite comprising only about 10\% of vehicles on the road \citep{intro_trucking_leslie}. Against this background, resource substitution among multiple, potentially composite and interchangeable, resource types emerges as an efficient solution, creating a win–win–win opportunity: it enables timely product delivery, supports real-time operational decisions, and reduces carbon impact. When compared with empty resource moves, resource substitution is highly preferred since it incurs no additional operating cost.

This paper studies the resource substitution problem in large networks. It addresses two complexities: (i) the resource substitution can only happen among interchangeable resources, and (ii) a solution that maximizes global utility may not be acceptable because its implementation costs can be unfairly distributed among schedulers. Thus, this paper studies resource substitution under the \emph{fairness} lens. Real-world examples include fleet reassignment in airlines \citep{intro_airline_jarrah}, containers repositioning in shipping yards \citep{intro_shipping_lin}, and bike rebalancing \citep{intro_bikeBenchmark_dell}, all of which are shown to be NP-hard. Although optimal solutions to these substitution problems can theoretically be computed, practical execution frequently depends on distributed decision making by schedulers, complicating the implementation of centralized plans. This is particularly true when performance incentives are tracked at the scheduler level rather than globally.

This research is motivated by real-world business challenges. The need for substitution arises when the flow of products using resources (equipment) creates an imbalance within a large network from one week to another, mismatching the projected inventory levels. One way to measure equipment imbalance is to compute the difference between the inventory of each equipment type at each node at the start and end of the week. Then, a surplus or deficit (positive or negative change) in the inventory level of a specific equipment type at a specific node is called an imbalance, and the total imbalance can be assessed by summing the absolute values of the imbalances for all equipment types at all nodes over the given time horizon (e.g., a week). To tackle this imbalance, the resource substitution typically occurs weekly upon the release of the upcoming week’s schedule. This paper studies resource balancing in networks via resource substitution. The primary objective is to minimize total network imbalance. Secondary objectives include achieving this with as few resource substitutions as possible and ensuring fair recommendations that support decentralized execution of a centralized policy, benefiting all stakeholders (e.g., end-users and schedulers).

The main contributions of this paper are summarized as follows:

\begin{enumerate}
\item \textbf{A general framework for fair resource substitution in large networks.} The paper proposes a framework that integrates operations research (OR) and machine learning (ML) to generate a portfolio of substitution recommendations for decision-makers.
  
\item \textbf{An OR component that models fairness explicitly.} The OR component models the resource substitution with fairness as a metric, and solves the problem using OR techniques. It overcomes key limitations of standard mixed-integer programming (MIP) formulations for resource substitution. The fairness metric emphasizes equitable substitutions between decentralized schedulers of a centralized policy.
  
\item \textbf{An ML component that leverages historical data.} The ML component leverages historical data to capture latent trends and scheduler preferences, enabling intelligent exploration of the decision space under a fairness perspective. It dynamically identifies the top-$\kappa$ candidate resources for each arc, thereby reducing the search space of the OR model and improving both solution quality and computational efficiency.
  
\item \textbf{Application to a real-world large-scale network.} The framework is applied to one of the largest package delivery companies in the world, which motivated this study. Computational experiments on realistic instances demonstrate substantial benefits compared with existing substitution-based resource allocation methods \citep{yang2022substitution}. By integrating machine learning and operations research components, the framework reductes  problem size by over 80\%  while preserving optimality, alongside a 90\% decrease in execution time. This highlights the framework effectiveness in enhancing both computational efficiency and scalability in large-scale operational contexts.
  
\end{enumerate}

The remainder of this paper is structured as follows. Section \ref{section:2} discusses the literature review. Section \ref{section:3} presents the general framework. Section \ref{section:4} highlights the case study from the industrial partner, while Section \ref{section:5} concludes the paper.

\section{Literature Review}\label{section:2}

This section reviews the relevant literature in three parts: resource reallocation, fairness in optimization, and the integration of ML and OR.

\subsection{Resource Reallocation}

Resource reallocation, also referred to as resource repositioning or redistribution, has been extensively studied, particularly within the context of freight transportation. Early research by \citet{leddon1967scheduling}, \citet{misra1972linear}, and \citet{white1972dynamic} developed deterministic models for the repositioning of rail cars and containers under complete information about future demand. More recent studies have incorporated repositioning decisions jointly with load assignments to address larger-scale and more realistic operational settings \citep{abrache1999new,erera2005global}. However, deterministic approaches are highly dependent on the accuracy of demand forecasts, which limits their robustness in uncertain environments \citep{powell1986stochastic,powell1987operational,crainic1993dynamic,yang2022substitution}.

Repositioning has also been extensively examined in the bike-sharing literature, where spatially imbalanced flows between stations create persistent supply–demand mismatches. Prior studies have investigated efficient and cost-effective strategies for fleet management to meet anticipated demand \citep{nair2011fleet,shu2013models,bruglieri2014vehicle}. Dynamic repositioning in this context is often guided by heuristic or rule-based methods \citep{ghosh2017dynamic,li2018dynamic}. More recent contributions have introduced clustering-based approaches for efficient routing \citep{lv2020hybrid}, incentive-based rebalancing schemes that leverage user participation \citep{chung2018bike}, and advanced heuristics that outperform traditional MIP formulations in both scalability and solution quality \citep{schuijbroek2017inventory,freund2022minimizing}.

Empty resource repositioning strategies can generally be classified into decentralized and centralized models. In decentralized frameworks, individual facilities or depots make independent decisions to optimize local fleet performance. For instance, \citet{du1997fleet} developed a decentralized stock control policy for fleet sizing and repositioning within a terminal system, while \citet{song2005optimal} analyzed container arrivals and repositioning decisions in a two-depot setting. Conversely, centralized models optimize decisions across the entire service network to enhance system-wide performance metrics \citep{jansen2004operational}.

Because demand uncertainty critically influences repositioning efficiency, recent work has incorporated stochastic optimization techniques. \citet{erera2009robust} proposed robust strategies to hedge against uncertainty, and \citet{long2012sample} applied two-stage stochastic programming to manage variability in container demand and supply.

\subsection{Fairness}

Fairness has emerged as a major concern in contemporary decision-making problems \citep{lodi2024framework} and has gained growing attention within the OR community (for a recent survey, see \citealp{fair_Guide_ChenHooker}). Efforts have increasingly focused on translating fairness principles into practical optimization frameworks applicable to real-world contexts such as supply chains and scheduling \citep{fair_SocialWelfare_ChenHooker}.

\citet{fair_PriceOfFairness_bertsimas} introduced the Price of Fairness (PoF), which measures the efficiency loss incurred when fairness constraints are imposed on optimal solutions. Building on this idea, \citet{fair_InequityAverse_karsu} and \citet{fair_FacilityLocation_gupta} examined fairness through equity and balance metrics, offering models that explicitly trade off efficiency for more equitable outcomes. More recently, \citet{fair_2agents_yu} extended these concepts to two-agent scheduling problems, analyzing outcomes through the lenses of proportionality, envy-freeness, and social welfare.

However, fairness has received comparatively less attention in the context of equipment substitution and resource allocation. Decentralized models tend to promote local fairness by granting individual terminals greater control over their own resources. In contrast, centralized models, while often more efficient from a system-wide perspective, can inadvertently create local imbalances by optimizing solely for global objectives. \citet{xie2017empty} emphasized that this trade-off parallels coordination challenges commonly observed in distributed logistics and game-theoretic frameworks, particularly in intermodal transport systems.

\subsection{Machine Learning for Optimization}

ML has been increasingly adopted to enhance operations research (OR) methodologies (for recent surveys, see \citealp{ml_survey4opt_bengio,scavuzzo2024machine}). Although the literature spans overlapping categories, such as solver acceleration, heuristic decision-making, and learning-based optimization, there is broad consensus on the growing synergy between ML and OR. A central objective in this integration is to leverage learned policies and value functions to replace manually designed heuristics within optimization algorithms. By framing optimization problems as data instances, researchers aim to enable cross-instance generalization, allowing models trained on specific problem distributions to improve decision quality and computational efficiency across similar problem classes.

Recent studies have demonstrated the potential of ML to accelerate and enhance optimization processes. \citet{ml_aircrew_tahir} showed that neural networks trained on historical optimal or near-optimal solutions can improve both solution quality and computational time. In particular, neural architectures are used to predict which variables or constraints are most promising to include or exclude during the solution process, thereby guiding solver decisions. Similarly, \citet{ml_column_morabit} demonstrated that supervised learning can effectively guide column selection in column generation frameworks, allowing solvers to focus on the most promising variables. This concept was later extended by \citet{ml_arcselection_morabit} to arc selection in constrained shortest path problems, where ML models predicted correct arcs, substantially reducing the search space in complex scheduling and routing applications.

In the context of supply chain resilience, \citet{wu2024towards} proposed a generic and scalable re-optimization framework designed to address perturbations caused by global disruptions. Their deep neural network architecture reduces the number of complicating variables, achieving roughly 20\% improvement in computational efficiency. More recently, \citet{akhlaghi2025propel} introduced PROPEL, an integrated ML–OR framework for large-scale Supply Chain Planning (SCP). PROPEL employs a two-stage strategy: supervised learning is used to predict integer variables that can be fixed to zero, while deep reinforcement learning (DRL) selectively relaxes these decisions to enhance solution quality. This approach has shown remarkable computational benefits, achieving up to 88\% reduction in primal gap and a 13–15$\times$ improvement in performance on industrial-scale SCP problems involving millions of variables.

\subsection{Paper Positioning}

This paper investigates optimization strategies for centralized policy design combined with decentralized execution across multiple jurisdictions. Each jurisdiction, managed by a local network scheduler, comprises resource nodes that enable opportunities for inter-jurisdictional collaboration and coordination.

From a fairness perspective, this study extends substitution-based approaches in the literature \citep{yang2022substitution} by explicitly integrating fairness considerations into both the OR and ML components. The objective is to develop resource reallocation plans that are not only computationally efficient but also equitable across locations and fair to individual decision makers involved in decentralized execution.

While prior studies in related domains, such as airline planning, where the top-$\kappa$ flights following a given flight are considered \citep{quesnel2022deep}, and vessel assignment, where the top-$\kappa$ quays are evaluated for each vessel \citep{wu2024towards}, have relied on static values of $\kappa$, their static approaches do not account for the structural context of individual arcs within the network. This work introduces a dynamic top-$\kappa_a$ mechanism for each arc $a \in \mathcal{A}$, in which $\kappa_a$ is adapted based on the betweenness centrality of each arc \citep{newman2018networks}, enabling a context-sensitive reduction of the search space. Thus, $\kappa$ varies across arcs according to their structural characteristics, quantified through an arc betweenness metric that reflects their relative importance in the network.

The proposed dynamic top-$\kappa$ approach represents a novel contribution not only to the resource substitution literature but also to the broader class of optimization problems that couple operations research with data-driven learning. By adapting $\kappa$ based on network structure and contextual features, this mechanism introduces a generalizable principle for adaptive search-space reduction in hybrid OR–ML frameworks, with potential applicability to other domains such as airline scheduling, maritime logistics, and transportation planning.

\section{General Framework}\label{section:3}

This section introduces the proposed fairness framework before detailing the OR component, the ML component, and the solution portfolio.

\subsection{Fairness Framework Description}

\begin{figure}[t!]
    \centering
    \includegraphics[width=\linewidth]{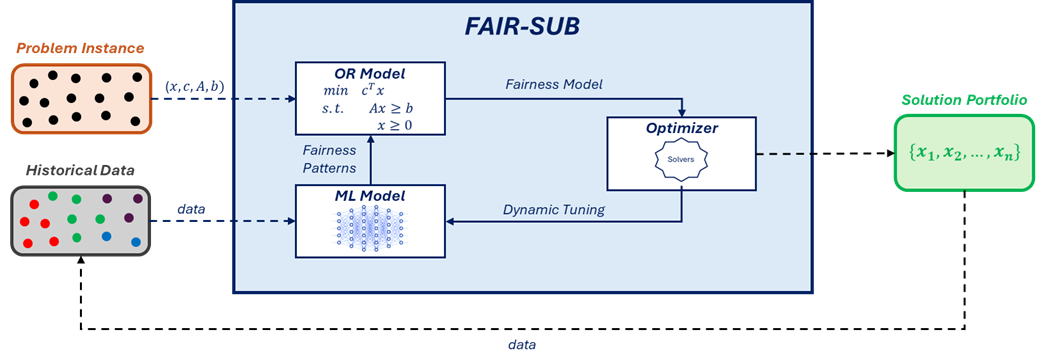}
    \caption{The Fairness Framework.}
    \label{fig:labeling}
\end{figure}

The proposed fairness framework for resource substitution, referred to as \emph{FAIR-SUB} and depicted in Figure \ref{fig:labeling}, takes as input both the problem instance and relevant historical data, and outputs a solution portfolio for local scheduler(s). The problem instance serves as the basis for constructing the OR model, while the historical data is used by the ML model to extract insights that support the OR formulation. The strengths of both models are integrated through an \emph{optimizer}, which represents the collection of computational techniques used to tackle large-scale optimization problems \citep{erraqabi5}. These include metaheuristics \citep{erraqabi1}, exact algorithms \citep{erraqabi2}, and commercial or open-source solvers \citep{himmich2023mpils}.

A key aspect of the FAIR-SUB framework is that both the OR and ML components explicitly account for fairness. In the OR model, fairness is embedded in the objective function through established metrics from the literature \citep{fair_Guide_ChenHooker}. In parallel, the ML model implicitly captures fairness by learning from historical scheduler preferences and patterns, ensuring that the resulting recommendations balance efficiency and fairness. The optimizer feedback can be used to tune the ML model dynamically. The solution portfolio is added to the historical data.

\subsection{The OR Component}

The resource substitution problem is formally defined as follows. Consider a directed graph $\mathcal{G}=(\mathcal{N},\mathcal{A})$, where $\mathcal{N}$ is the set of nodes and $\mathcal{A}$ is the set of arcs. Each arc $a \in \mathcal{A}$ corresponds to a task to be performed by an allocated resource $r \in \mathcal{R}$. When the available resources are performing the tasks (e.g., flow through the network) over a given planning horizon, an imbalance might occur within each node. Let $A^{+}_n=\{(i,n): i \in \mathcal{N}\}$ (resp. $A^{-}_n=\{(n,i): i \in \mathcal{N}\}$) the set of incoming (resp. outgoing) arcs at node $n \in \mathcal{N}$. A resource assignment is a function $\Phi: \mathcal{A} \rightarrow \mathcal{R}$ that assigns a resource to each arc and the initial resource assignment is denoted $\Phi_0$. The imbalance of an assignment $\Phi$ can be computed as follows:
\[ \mathcal{I}(\Phi):=\sum_{r\in\mathcal {R}}\sum_{n \in \mathcal{N}} \; |\sum_{a \in A^{+}_{nr}} \mathbbm{1} (\Phi_0(a)=r) - \sum_{a \in A^{-}_{nr}} \mathbbm{1}(\Phi_0(a)=r) \, |.
\]
\emph{FAIR-SUB} tackles the imbalance problem by substituting resources within the arcs of network. It consists of two stages. Stage 1 computes the minimal imbalance $I^*$ of the network. An assignment $\Phi$ is optimal when $\mathcal{I}(\Phi)= I^*$. Stage 2 computes the optimal $\Phi^*$ that is closest to $\Phi_0$, i.e., $\Phi^*\in \arg \min_{\mathcal{I}(\Phi)=\mathcal{I}^*} ||\Phi-\Phi_0||$, where $||\Phi-\Phi_0||=|\{a \in \mathcal{G}: \Phi(a)\neq\Phi_0(a)\}|$.

Denote by $\mathcal{R}_a$ the set of resources that can be used on arc $a \in \mathcal{A}$ (i.e., the resources that can perform the corresponding task). The decision variables of Stage 1 are
\begin{equation}
    x_{ar}=\begin{cases}
        1, \hspace{2mm} \text{if resource $r$ is used on arc $a$,}  \\
        0, \hspace{2mm} \text{otherwise.}
    \end{cases}
\end{equation}
and $I_{nr} \geq 0$ captures the imbalance of resource $r$ at node $n$. The Stage 1 model is defined as follows:

\begingroup
\allowdisplaybreaks
\begin{align}
\label{Stage 1 Imbalance}
I^*=\min \hspace{1mm} & \sum_{n \in \mathcal{N}}\sum_{r \in \mathcal{R}} I_{nr} &&\\
s.t. \hspace{1mm} \label{Stage 1 Constraints 1} & \sum_{a \in \mathcal{A}^+_n} x_{ar} - \sum_{a \in \mathcal{A}^-_n} x_{ar} \leq I_{nr}, && \forall n \in \mathcal{N},\\
\hspace{1mm} \label{Stage 1 Constraints 2} & \sum_{a \in \mathcal{A}^-_n} x_{ar} - \sum_{a \in \mathcal{A}^+_n} x_{ar} \leq I_{nr}, && \forall n \in \mathcal{N},\\
\hspace{1mm} \label{Stage 1 Constraints 3} & \sum_{r \in \mathcal{R}_a} x_{ar} = 1, && \forall a \in \mathcal{A},\\
\label{Stage 1 Constraints 4} & x_{ar} \in \{0,1\}, && \forall a \in \mathcal{A}, r \in \mathcal{R}
\end{align}
\endgroup

Objective (\ref{Stage 1 Imbalance}) minimizes the total imbalance. Constraints (\ref{Stage 1 Constraints 1}) and (\ref{Stage 1 Constraints 2}) ensure that the imbalance within each node for each resource is equal to the net surplus or deficit of that resource at that node. Constraints (\ref{Stage 1 Constraints 3}) ensure the assignment of exactly one resource to each arc.

The Stage 2 model minimizes the number of changes $\Delta^*$, and with Constraints (\ref{Stage 2 Constraints 1}), ensures the obtained resource substitution is optimal:

\begin{align}
\label{Stage 2 Imbalance}
\Delta^*=\min \hspace{1mm} & \sum_{a \in \mathcal{A}}(1-x_{a\Phi_0(a)}) &&\\
s.t. \hspace{1mm} \label{Stage 2 Constraints 1} & \sum_{n \in \mathcal{N}}\sum_{r \in \mathcal{R}} I_{nr} \leq I^*, && \\
\hspace{1mm} \notag & (\ref{Stage 1 Constraints 1}) \hspace{1mm} \text{to} \hspace{1mm} (\ref{Stage 1 Constraints 4}). &&
\end{align}

While this approach allows for finding the minimal imbalance $I^*$ and the minimal number of changes $\Delta^*$ to bring balance back to the network, it might not be fair for all the schedulers involved. A \emph{scheduler} is a decision maker in charge of a set of nodes within the network. Some schedulers may be required to implement disproportionately more changes. Some manage large or high-throughput areas; others oversee fewer nodes with intense volume. Coordination practices also vary: some collaborate regularly with neighbors, while others operate in relative isolation. Formally, let $\mathcal{S}$ be the set of schedulers for network $\mathcal{G}$, and $\mathcal{N}_s$ be the set of nodes corresponding to scheduler $s \in \mathcal{S}$. To incorporate fairness, the objective function in Stage 2 is modified to reflect one of the fairness metrics proposed in the literature \citep{fair_Guide_ChenHooker}: the \emph{Minimax} metric that minimizes the maximum changes per schedule. Let $A_s=\{(i,j): i,j \in \mathcal{N}_s\} \cup \{(i,j): i \in \mathcal{N}_s, j \notin \mathcal{N}_s\}$, i.e., the arcs corresponding to scheduler $s$ are the arcs between their nodes and the arcs outgoing from one of their nodes to nodes assigned to different schedulers. Stage 2 model becomes

\begingroup
\allowdisplaybreaks
\begin{align}
\label{Fair Stage 2 Imbalance}
\Delta^{fair}=\min & \max_{s \in \mathcal{S}} \sum_{a \in \mathcal{A}_s}(1-x_{a\Phi_0(a)}) &&\\
s.t. \hspace{1mm} \label{Fair Stage 2 Constraints 1} & \sum_{n \in \mathcal{N}}\sum_{r \in \mathcal{R}} I_{nr} \leq I^*, && \\
\hspace{1mm} \notag & (\ref{Stage 1 Constraints 1}) \hspace{1mm} \text{to} \hspace{1mm} (\ref{Stage 1 Constraints 4}). &&
\end{align}
\endgroup

This model can be linearized as follows

\begin{align}
\label{Linear Fair Stage 2 Imbalance}
\Delta^{fair}=\min \hspace{1mm} & Z &&\\
s.t. \hspace{1mm} \label{Linear Fair Stage 2 Constraints 1} & Z \geq \sum_{a \in \mathcal{A}_s}(1-x_{a\Phi_0(a)}), && \forall s \in \mathcal{S},\\
\hspace{1mm} \notag & (\ref{Stage 1 Constraints 1}) \hspace{1mm} \text{to} \hspace{1mm} (\ref{Stage 1 Constraints 4}), (\ref{Fair Stage 2 Constraints 1}). &&
\end{align}

\noindent 
The following example compares the efficient and fair models.

\begin{exam}
Figure~\ref{Example 1} presents a scenario with six resources ($r_1$–$r_6$) and three schedulers ($s_1$–$s_3$), each operating three nodes. As shown in Figure~\ref{fig:example_Original}, the initial imbalance is 20, while the minimal achievable imbalance is 12. To illustrate the reallocation required to attain this minimal imbalance, Figures~\ref{fig:example_EJOR} and~\ref{fig:example_FAIR} depict two scenarios corresponding to Models~(\ref{Stage 2 Imbalance}) and~(\ref{Fair Stage 2 Imbalance}), respectively. Changes relative to the original configuration are highlighted in purple.      

\begin{figure}[t!]
    \centering
    \begin{subfigure}[b]{0.3\textwidth}
        \includegraphics[width=\textwidth]{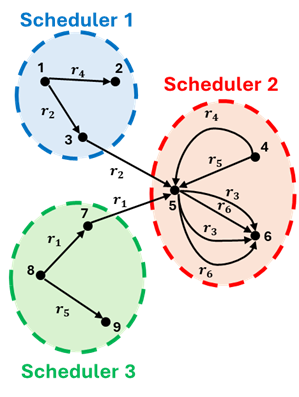}
        \caption{The Original Scenario.}
        \label{fig:example_Original}
    \end{subfigure}
    \hfill
    \begin{subfigure}[b]{0.3\textwidth}
        \includegraphics[width=\textwidth]{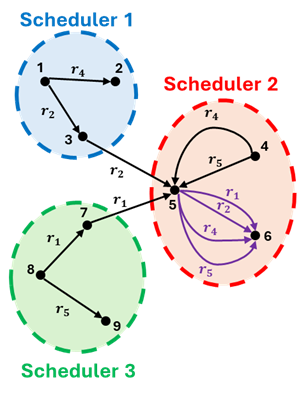}
        \caption{The Efficient Scenario.}
        \label{fig:example_EJOR}
    \end{subfigure}
    \hfill
    \begin{subfigure}[b]{0.3\textwidth}
        \includegraphics[width=\textwidth]{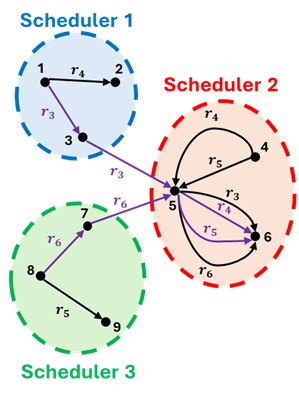}
        \caption{The Fair Scenario.}
        \label{fig:example_FAIR}
    \end{subfigure}
    \caption{Example with 6 Resources and 3 Schedulers operating 3 Nodes Each.}
    \label{Example 1}
\end{figure}

When considering efficiency, the model reaches the minimal imbalance by substituting 4 resources on the arcs corresponding to scheduler 2. No changes are required for the other two schedulers. When considering fairness, the model reaches the minimal imbalance by substituting 2 resources on the arcs corresponding to each of the three schedulers. In a collaborative environment, the fair scenario is more implementable since all schedulers have 2 substitutions each, compared to only one scheduler having 4 substitutions.
\end{exam}

The OR models above focus on two primary objectives: (i) minimizing network imbalance in Stage 1 and (ii) limiting the number of resource changes in Stage 2. These quantities serve as the operational “costs’’ relevant to the decision-making setting studied. Without loss of generality, this model considers the number of changes and the total network imbalance as the only cost components, which allows us to isolate and analyze the structural properties of the resource substitution problem.

That said, the proposed framework is flexible and can accommodate richer cost structures. In particular, arc-dependent substitution costs, such as the cost of deploying a super-long resource instead of a short one, or distance-based penalties along the arc, can be incorporated directly into the objective or as additional constraints.

\subsection{The ML Component}\label{ML Component}

Using historical operational data collected by companies, it is possible to develop ML models that learn schedulers’ preferences \citep{ml_survey4opt_bengio}, thereby promoting fairness in decision making. Such models help uncover latent patterns and behavioral trends that can enhance both the OR model and the optimizer’s performance. To construct these ML models, it is essential to specify the target variable, define the historical data structure, identify relevant features, and determine the network architecture. While numerous machine learning techniques apply to this task, the following approach is designed to address the specific structure and requirements of the resource substitution problem.

\noindent \textbf{Target.} In the resource substitution framework, the ML component aims at estimating the likelihood that a specific resource $r \in \mathcal{R}$ be assigned to an arc $a \in \mathcal{A}$.  Formally, the model predicts probabilities $p_{ar}$ representing the propensity of resource $r$ to fulfill the task associated with arc $a$. These estimates are then used to identify the most promising candidates by selecting the top $\kappa$ resources for each arc, where $\kappa$ denotes an integer cut-off parameter. By focusing only on these high-probability resource–arc pairs, the ML component effectively reduces the decision space of the optimization models in Stage~1 (Model~\ref{Stage 1 Imbalance}) and Stage~2 (Model~\ref{Fair Stage 2 Imbalance}), thereby decreasing the number of binary decision variables $x_{ar}$ to be considered. For each arc $a$, the model produces a probability vector $Y_a \in \mathbb{R}^{|\mathcal{R}|}$, where the $r$-th entry corresponds to $p_{ar}$. Sorting $Y_a$ in descending order of $p_{ar}$ enables the extraction of the top-$\kappa$ resources, which are subsequently used as the candidate set for substitution decisions.

Unlike prior approaches that adopt a fixed, static $\kappa$ across all arcs, this study introduces a dynamic top-$\kappa_a$ mechanism, in which $\kappa_a$ is determined adaptively based on each arc’s betweenness centrality \citep{newman2018networks}. Arcs with high betweenness, indicating strong structural importance and greater potential for flow interaction, are assigned larger $\kappa_a$ values, allowing for broader exploration of substitution options. Conversely, arcs with low betweenness receive smaller $\kappa_a$ values, emphasizing computational efficiency in regions where local changes have limited impact. This dynamic treatment of $\kappa_a$ ensures that the reduction of the search space is both context-aware and structurally informed, enhancing scalability while maintaining fairness and solution optimality.

The structural importance of each arc $a \in \mathcal{A}$ is quantified using its \emph{betweenness centrality}, denoted by $B(a)$, and specified as
\begin{equation}
B(a) = \sum_{\substack{i,j \in \mathcal{N} \\ i \neq j}} \frac{\sigma_{ij}(a)}{\sigma_{ij}},
\end{equation}
where $\sigma_{ij}$ is the total number of shortest paths between nodes $i$ and $j$, and $\sigma_{ij}(a)$ is the number of those paths that pass through arc $a$. A higher $B(a)$ indicates that arc $a$ is more frequently used as a connector between node pairs, reflecting its structural importance.

Based on their betweenness values, arcs are categorized into three classes:
\begin{enumerate}
    \item \textbf{Low-betweenness arcs} ($B(a) < \tau_1$): peripheral connections;
    \item \textbf{Medium-betweenness arcs} ($\tau_1 \le B(a) < \tau_2$): moderately connected links;
    \item \textbf{High-betweenness arcs} ($B(a) \ge \tau_2$): major connectors or bottlenecks in the network.
\end{enumerate}
These classes inform the assignment of dynamic top-$\kappa_a$ values. Specifically, high-betweenness arcs are allocated larger $\kappa_a$ values to allow for broader exploration of substitution options, while low-betweenness arcs receive smaller $\kappa_a$.

\noindent \textbf{Data \& Features.} The data used for training the ML model depends on the specific application context. Once the historical data is identified, appropriate features are selected to construct the feature vector $X_{ar}$ for each arc $a \in \mathcal{A}$ and resource $r \in \mathcal{R}$. In general, features may include network-related attributes (e.g., origin and destination nodes), operational responsibilities (e.g., schedulers associated with each node), and task characteristics (e.g., quantity, time of day, duration).

\noindent \textbf{Network Architecture.} Given the richness of the available data, each observation is represented by a high-dimensional feature vector. In such settings, individual features often exhibit weak correlation with the prediction target, implying that accurate forecasting requires a model capable of capturing complex, nonlinear interactions among features. Deep neural networks (DNNs) are particularly well-suited to this task due to their ability to learn hierarchical representations from data \citep{goodfellow2016deep,lecun2015deep}. A DNN consists of multiple layers of interconnected computational units, or neurons. The input layer encodes the explanatory features, while the output layer produces the predicted probability $p_{ar}$ associated with assigning resource $r \in \mathcal{R}$ to arc $a \in \mathcal{A}$. Between these layers, one or more hidden layers enable the model to capture intricate relationships between the inputs and outputs. Each neuron applies a weighted linear combination of its inputs, followed by a nonlinear activation function, which allows the network to approximate complex functional mappings. During the training phase, the model iteratively adjusts its weights and activation parameters to minimize prediction error over the training dataset. Once trained, the DNN can estimate $p_{ar}$ for unseen data instances, providing a scalable and flexible mechanism for predicting resource–arc preferences across large transportation networks.

\subsection{Solution Portfolio}

Once the OR and ML models are designed, it is possible to combine the ML insights generated from the historical data with the OR models to boost the optimizer's performance in solving the OR models. While the solving might be quick and optimal, it might also take more time than expected, especially in large-scale contexts, which require additional computational strategies. In such contexts, it is advisable to employ established mathematical optimization strategies from the literature, including metaheuristics, decomposition approaches, and parameter tuning of the underlying MIP solver. Examples include: (1) the iterated local search metaheuristic, which starts from the initial assignment $\Phi_0$ and explores solutions around it locally to find resource substitutions that decrease the imbalance \citep{lourencco2003iterated}; (2) the Dantzig-Wolfe decomposition, which starts with an initial set of $x_{ar}$ variables and uses column generation to identify and add more promising $x_{ar}$ variables \citep{desaulniers2006column}; (3) the parameter configuration, which consists of using available tuners in the literature to identify better configurations than the default configuration of the considered solver on the given problem instance \citep{lopez2016irace,himmich2023mpils}.

Although only two formulations are illustrated here (the efficient and minimax models), the same framework can be extended to evaluate numerous formulations under different fairness metrics.\citep{fair_Guide_ChenHooker}. After solving the problem using the various formulations, the suggestion is to provide the schedulers with a portfolio of solutions from which they can choose. This portfolio contains a set of Pareto-optimal solutions that might not be optimal from a solver perspective, but ensure a fair resource substitution in the considered network $\mathcal{G}$ \citep{fair_PriceOfFairness_bertsimas}.

Beyond offering a set of Pareto-efficient trade-offs, the solution portfolio also provides interpretable and actionable recommendations for the schedulers. Each substitution suggested within the portfolio can be classified as either an \emph{internal substitution} or a \emph{collaborative substitution}. Internal substitutions occur between two nodes managed by the same scheduler and reflect opportunities for local reallocation of resources within a single operational jurisdiction. In contrast, collaborative substitutions involve arcs linking nodes managed by different schedulers, thereby promoting cross-scheduler coordination and network-level efficiency. This classification allows decision makers to understand whether improvements arise from local flexibility or inter-scheduler collaboration, enabling more informed and context-aware decision support in real-world scheduling environments.

\section{Package Delivery Company Case Study}\label{section:4}

This section presents a case study conducted with a package delivery company. The operational context is described first, followed by a detailed explanation of the experimental design used to evaluate the proposed framework and the corresponding computational results. The section concludes with a discussion of the managerial insights derived from the analysis, emphasizing the implications for both operational efficiency and fairness in resource allocation.

\subsection{Problem Description}

The case study investigated in this paper focuses on a resource substitution problem encountered by a major package delivery company. In this context, packages are transported between facilities and customers through a large network of interconnected terminals. Each week, the company must allocate and position equipment across facilities to meet anticipated package flows for the upcoming period. This planning task is complicated by network imbalances, forcing schedulers to reposition equipment: an empty, costly, and non-productive operation. Efficient resource allocation and substitution decisions play a crucial role in ensuring on-time deliveries, supporting operational decision making, and minimizing the environmental footprint associated with empty movements.

One approach to measuring equipment imbalance is to consider differences in the inventory of different equipment types at a facility at the start of the week and at the end of the week. A surplus or deficit (positive or negative change) in the inventory level of a specific equipment type at a specific terminal is called an imbalance, and the total imbalance of a plan can be assessed by summing the absolute values of the imbalances for all equipment types at all terminals. Equipment substitution complements the empty repositioning of equipment, but it is only possible if companies operate multiple, interchangeable equipment types. When compared with repositioning empty equipment, equipment substitution may be preferred since this may not require any additional operating cost. The company operates several equipment types, which can be grouped into 3 groups: trailers with a length of 28 feet called shorts, trailers with a length ranging from 40 to 48 feet called longs, and trailers with a length of 53 feet called extra-longs. These can be composite and are mostly interchangeable (see Appendix \ref{A} where all the substitution matrix for all equipment types is given).

The balancing process occurs weekly with the release of the schedule for the upcoming week. Schedulers aim at minimizing net imbalances within their facilities (nodes), often referred to as “one in, one out.” Schedulers may be responsible for various facility types: hubs (large and centralized facilities that handle high volumes of shipments and serve as major distribution points in the network), centers (mid-sized facilities that coordinate regional operations), and locals (small facilities focused on last-mile delivery and direct customer interactions within a limited geographic area). A given jurisdiction may contain up to 90 facilities, with cooperation between connected jurisdictions. Connectivity between each pair of nodes is not guaranteed.

\subsection{Experimental design}

This section describes the test instances and the ML model used in the experiments. Both the OR and ML components are implemented in Python, and optimization experiments are solved using Gurobi version 12.0.2. All computations were performed on a 40-core machine with 384 GB of RAM, running Oracle Linux Server 7.7 on a 3.20 GHz Intel® Core™ i7-8700 processor. The ML models are implemented using the PyTorch library. For evaluation, real-time runtime is reported to assess computational performance across all test instances.

\subsubsection{Instances}

The case study is based on eight classes of real-world problem instances supplied by one of the largest package delivery companies in the world. Key characteristics of these instances include the number of schedulers, nodes, arcs, equipment, and the initial network imbalance. For each class, 10 instances are generated, resulting in a total of 80 test cases. The average values of these features for each class are summarized in Table \ref{Instances}.

\begin{table}[t!]
  \centering
  \caption{Instances}
  \label{Instances}
    \begin{tabular}{c|c|c|c|c|c}
    \toprule
    \textbf{Class} & \textbf{\# Sched.} & \textbf{\# Nodes} & \textbf{\# Arcs} & \textbf{\# Equip.} & \textbf{$I_0$} \\
    \midrule
    1     & 2     & 167    & 138992 & 13    & 4236 \\
    \midrule
    2     & 5     & 356   & 509557 & 16    & 10644 \\
    \midrule
    3     & 8     & 536   & 893028 & 17    & 18370 \\
    \midrule
    4     & 11    & 612   & 1278907 & 18    & 22733 \\
    \midrule
    5     & 15    & 709   & 1816934 & 19    & 28879 \\
    \midrule
    6     & 20    & 974   & 3004375 & 20    & 40799 \\
    \midrule
    7     & 30    & 1296   & 5259231 & 22    & 58630 \\
    \midrule
    8     & 40    & 1588   & 8341357 & 23    & 78488 \\
  \bottomrule
  \end{tabular}
\end{table}

For brevity, only average results for each class are reported, while instance-level results are provided in the Online Supplement.

\subsubsection{Machine Learning Model}

The deep neural network (DNN) is trained to predict the most probable equipment type associated with each origin–destination (O–D) movement. As described in Section~\ref{section:3}, the ML component constructs a probability vector $Y_a$ for every arc $a \in \mathcal{A}$, where each entry corresponds to the likelihood of assigning a specific resource $r \in \mathcal{R}$. The model is developed within a supervised learning framework using data derived from previously implemented solutions, hereafter referred to as \emph{reference solutions}. Each reference solution $\Bar{x}$ specifies one resource assignment per arc, thereby yielding $|A_{\Bar{x}}|$ training instances of the form $(X_{ar}, y_{ar})$, with $a \in \mathcal{A}_{\Bar{x}}$ and $r \in \mathcal{R}$. To enhance learning robustness, all equivalent optimal reference solutions are included in the training set by exploiting problem symmetries. This augmentation allows for multiple feasible assignments to be represented for a given arc $a$, capturing variability in optimal or near-optimal configurations. For each pair $(a, r)$, the associated label $y_{ar}$ reflects the empirical frequency with which resource $r$ is assigned to arc $a$ across all reference solutions, thus encoding observed assignment patterns as probabilistic targets for model training.

Each arc $a \in  \mathcal{A}_{\Bar{x}}$ and resource $r \in \mathcal{R}$ have a feature vector $X_{ar}$, which contains the following features: 
\begin{enumerate}
    \item A categorical feature indicating the origin node, denoted $f^{(1)}$.
    \item A categorical feature indicating the destination node, denoted $f^{(2)}$.
    \item A numerical feature indicating the volume to be transported, denoted $f^{(3)}$.
    \item A categorical feature indicating the scheduler in charge of the origin node, denoted $f^{(4)}$.
    \item A categorical feature indicating the scheduler in charge of the destination node, denoted $f^{(5)}$.    
    \item A numerical feature corresponding to the time of the day, denoted $f^{(6)}$.
    \item A numerical feature corresponding to the time of the week, denoted $f^{(7)}$.
    \item A numerical feature corresponding to the miles between the origin and destination nodes, denoted $f^{(8)}$.
    \item A categorical feature corresponding to the size of the equipment required to transport the volume, denoted $f^{(9)}$.
\end{enumerate}


The DNN is trained using a pool of reference solutions compiled over a 13-week historical horizon. For each candidate pair $(a, r)$, the model learns to predict a label corresponding to the likelihood of assigning resource $r$ to arc $a$. The predicted label for each input vector $X_{ar}$ is denoted by $y_{ar}^{\text{prd}}$. To evaluate model generalization, the dataset of labeled pairs $(X_{ar}, y_{ar})$ is randomly partitioned into three mutually exclusive subsets: 60\% for training, 20\% for validation, and 20\% for testing. The split is stratified by equipment class to ensure that the distribution of target labels is consistent across all subsets, thereby avoiding biased performance estimates for underrepresented classes. Once trained, the model produces a predicted probability $y_{ar}^{\text{prd}}$ for each resource–arc pair. These predictions are then ranked in descending order to identify the most promising resource candidates for each arc. The resulting ranking enables the selection of the top-$\kappa$ resources, which are subsequently passed to the OR component for substitution optimization.

The machine learning architecture employed is a fully connected feedforward DNN. Consistent with standard design principles, the number of neurons in each hidden layer is set to be less than or equal to that of the preceding layer, ensuring a gradually contracting structure. All hidden-layer neurons use the rectified linear unit (ReLU) activation function to introduce nonlinearity, while the output layer consists of a single sigmoid neuron, producing a probability value bounded within the interval $[0,1]$. The model hyperparameters, including the number of hidden layers, neurons per layer, learning rate, and batch size, along with the training algorithm, are tuned via a randomized grid search (see Table~\ref{tab:ml_hyperparameters}). The final network configuration is selected based on its performance on the validation set.

\begin{table}[t!]
\caption{ML Model Hyperparameters and Training Configuration}
\label{tab:ml_hyperparameters}
\centering
\small
\renewcommand{\arraystretch}{1.1}
\setlength{\tabcolsep}{5pt}
\begin{adjustbox}{width=0.9\textwidth, center}
\begin{tabular}{lll}
\hline
\textbf{Category} & \textbf{Parameter} & \textbf{Value} \\
\hline
\textbf{Model Architecture} 
& Hidden Layers & 2 \\
& Neurons (per layer) & [128, 64] \\
& Output Neurons & 30 \\
& Activation (Hidden/Output) & ReLU/Softmax \\
\textbf{Training Configuration} 
& Optimizer & Adam \\
& Learning Rate & 0.001 \\
& Dropout Rate & 0.3 \\
& Loss Function & Categorical Crossentropy \\
& Epochs/Batch Size & 50/32 \\
\textbf{Data Splitting} 
& Train/Test/Validation Ratio & 60/20/20 \\
& Stratification & Yes (by resource type) \\
\textbf{Evaluation Metrics} 
& Primary/Secondary & Loss/AUROC \\
\hline
\end{tabular}
\end{adjustbox}
\end{table}

The predictive performance of the ML model is evaluated by assessing how well the top-ranked $\kappa$ resources correspond to the true labels across all arcs. Specifically, for each arc $a \in \mathcal{A}$, let $j \in {1,2,\ldots,\kappa}$ denote the index of the $j^{th}$-ranked resource in the ordered prediction vector $Y_a$, where $y_{a{r_1}}$ represents the highest predicted probability and $y_{a{r_\kappa}}$ the $\kappa^{th}$. The model effectiveness is then quantified using the \emph{label sum ratio}, computed as the proportion of true labels captured within the top-$\kappa$ predictions. This measure reflects the model ability to rank the most relevant resource assignments at the top of the list, providing a direct and interpretable indicator of ranking accuracy.

\begin{center}
$TOP_{\kappa} = \dfrac {\sum_{a \in \mathcal{A}}{\sum_{j=1}^{\kappa}} {y_{a{r_j}}}} {\sum_{a \in \mathcal{A}}{\sum_{r \in \mathcal{R}}} {y_{ar}}}$    
\end{center}

The DNN is trained in a supervised setting using the Adam optimizer \citep{kingma2014adam}, and multiple regularization mechanisms are incorporated to mitigate overfitting. Specifically, dropout is applied to hidden layers with probability 0.3. In addition, model performance, measured by the validation $TOP_\kappa$ metric, is monitored every ten epochs. Training is halted early if validation performance declines for two consecutive evaluations or upon reaching a predefined maximum number of epochs.

The training process for the neural network required less than two hours. In practical applications, this computational cost is negligible, as each ML model is trained once and subsequently deployed for extended periods (e.g., several months or years). Moreover, retraining the model is substantially faster—typically a few minutes—since the network parameters start from a near-optimal configuration and only require fine-tuning. The performance of the trained model is summarized in Table \ref{Average TOP}. Each arc can correspond to multiple potential equipment types, and the network performance is evaluated for $\kappa$ values ranging from 1 to 5. As discussed in Section \ref{ML Component}, the choice of $\kappa$ is guided by arc betweenness values, with a maximum value of 5.

\begin{table}[!t]
  \caption{Average $TOP_\kappa$ for the Test Pairs}
  \label{Average TOP}
  \centering
  \renewcommand{\arraystretch}{1.5}
  \begin{adjustbox}{width=0.65\textwidth, center}
  \begin{tabular}{cccccc}
    \hline
    $\kappa$ & 1 & 2 & 3 & 4 & 5 \\
    \hline
    Average $TOP_\kappa$ & 69.50\% & 87.20\% & 97.33\% & 99.10\% & 100.00\% \\
    \hline
  \end{tabular}
  \end{adjustbox}
\end{table}

The proposed ML model is employed to enhance both stages of the resource substitution framework. Unlike \citet{yang2022substitution}, who rely on predefined compatibility matrices similar to the one presented in Appendix \ref{A}, the proposed ML approach inherently satisfies compatibility constraints. Specifically, since the model is trained on historical data that only includes compatible equipment–arc pairings, it ensures that an arc’s initial equipment is substituted only with compatible alternatives. This data-driven formulation not only preserves operational feasibility but also leads to a substantial reduction in the number of arcs considered.

To prioritize substitutions where they have the greatest impact, directed edge betweenness centrality is computed on the network using a sampling-based approximation of $k$ for scalability, and each movement is classified as High, Medium, or Low based on the tuned betweenness thresholds $\tau_1$ and $\tau_2$. These thresholds were determined empirically through preliminary experiments. These exploratory runs were conducted on representative instances to identify breakpoints that yield balanced class distributions and stable performance across test cases. Next, the dynamic-$\kappa$ mechanism is applied, where movements on high-betweenness corridors receive a larger $\kappa$ value, movements on medium-betweenness corridors receive a medium $\kappa$ value, and low-betweenness movements are restricted to lower $\kappa$ values.

\subsection{Computational Performance}

This section evaluates the computational performance of the proposed framework across the eight problem classes introduced earlier. The experiments are designed to assess both the quality of the solutions and the efficiency gains achieved through the integration of the OR and ML components, particularly the dynamic top-$\kappa_{a}$ mechanism.

Table \ref{tab:Minimal Imbalance from Stage 1} summarizes the Stage 1 results (minimal imbalance) for the eight classes. In addition to the already presented features, Table \ref{tab:Minimal Imbalance from Stage 1} reports the number of internal arcs (i.e., the arcs linking nodes for which the scheduler is the same), the number of collaborative arcs (i.e., arcs for the scheduler is not the same), the minimal imbalance ($I^*$), the reduction in imbalance (Gain), which is computed as $\frac{I_0-I^*}{I_0}$, and the execution time (Time) in seconds.

\begin{table}[t!]
  \centering
  \scalebox{0.8}{
  \caption{Minimal Imbalance from Stage 1}
    \label{tab:Minimal Imbalance from Stage 1}
    \begin{tabular}{c|c|c|c|c|c|c|c|c|c|c}
    \toprule
    \textbf{Class} & \textbf{\# Sched.} & \textbf{\# Nodes} & \textbf{\# Arcs} & \textbf{\# Equip.} & \textbf{\# Inter.} & \textbf{\# Collab.} & \textbf{$I_0$} & \textbf{$I^*$} & \textbf{Gain} & \textbf{Time} \\
    \midrule
    1 & 2 & 167 & 138992 & 13 & 138709 & 283 & 4236 & 3444 & 18.69\% & 1.02 \\
    \midrule
    2 & 5 & 356 & 509557 & 16 & 462500 & 47056 & 10644 & 8357 & 21.49\% & 3.13 \\
    \midrule
    3 & 8 & 536 & 893028 & 17 & 811401 & 81627 & 18370 & 14709 & 19.93\% & 5.53 \\
    \midrule
    4 & 11 & 612 & 1278907 & 18 & 1089299 & 189608 & 22733 & 17667 & 22.28\% & 7.61 \\
     \midrule
    5 & 15 & 709 & 1816934 & 19 & 1448200 & 368734 & 28879 & 22863 & 20.83\% & 9.57 \\
     \midrule
    6 & 20 & 974 & 3004375 & 20 & 2213879 & 790496 & 40799 & 32641 & 20.00\% & 13.19 \\
      \midrule
    7 & 30 & 1296 & 5259231 & 22 & 3516181 & 1743050 & 58630 & 47863 & 18.37\% & 19.77 \\
      \midrule
    8 & 40 & 1588 & 8341357 & 23 & 4910272 & 3431084 & 78488 & 65655 & 16.35\% & 26.48 \\
    \bottomrule
    \end{tabular}
    }
\end{table}

Table \ref{tab:Minimal Imbalance from Stage 1} reveals that, as the size increases, so does the scale of the problem, which slightly impacts the computational effort, as seen in the longer execution times for Classes 7 and 8. Notably, the reduction percentage remains significant across all classes, indicating that the first stage reduces the imbalance regardless of problem size.

Table \ref{tab:Stage 2 for the Efficient and Fair Models} presents results comparing the efficient and fair Stage 2 models across the eight classes. In addition to the previously described features, the analysis reports the number of changes ($\Delta$), the maximum number of changes per scheduler ($Z$), and the execution time (Time) required by the second stage for both the efficient and fair models. The cost of fairness is defined as $\Delta^{FAIR} - \Delta^*$, while the gain in fairness is given by $Z^{FAIR} - Z^*$. The considered fairness metric is the \emph{minimax} metric. 

\begin{table}[t!]
  \centering
  \scalebox{0.8}{
  \caption{Stage 2 for the Efficient and Fair Models}
  \label{tab:Stage 2 for the Efficient and Fair Models}
    \begin{tabular}{c|c|c|c|c|c|c|c|c|c|c|c}
    \toprule
    \multirow{2}[4]{*}{\textbf{Class}} & \multirow{2}[4]{*}{\textbf{\# Sched.}} & \multirow{2}[4]{*}{\textbf{I$_0$}} & \multirow{2}[4]{*}{\textbf{I$^*$}} & \multicolumn{3}{c|}{\textbf{Efficiency}} & \multicolumn{3}{c|}{\textbf{Fairness}} & \multirow{2}[4]{*}{\textbf{$\Delta^{FAIR}-\Delta^*$}} & \multirow{2}[4]{*}{\textbf{$Z^{FAIR}-Z^*$}} \\
    \cline{5-10}
      &       &       &       & \textbf{$\Delta^*$} & \textbf{$Z^*$} & \textbf{Time} & \textbf{$\Delta^{FAIR}$} & \textbf{$Z^{FAIR}$} & \textbf{Time} &       &  \\
    \midrule
    1 & 2 & 4236 & 3444 & 321 & 199 & 1.44 & 357 & 188 & 1.53 & 35 & 11 \\
    \midrule 
    2 & 5 & 10644 & 8357 & 952 & 354 & 4.74 & 1135 & 242 & 5.50 & 183 & 112 \\
    \midrule 
    3 & 8 & 18370 & 14709 & 1542 & 452 & 8.27 & 1852 & 274 & 11.30 & 310 & 178 \\
    \midrule    
    4 & 11 & 22733 & 17667 & 2048 & 443 & 12.25 & 2796 & 285 & 15.50 & 748 & 159 \\
    \midrule    
    5 & 15 & 28879 & 22863 & 2701 & 486 & 15.14 & 4184 & 308 & 18.16 & 1483 & 178 \\
    \midrule    
    6 & 20 & 40799 & 32641 & 3725 & 710 & 22.13 & 7350 & 397 & 24.98 & 3624 & 313 \\
    \midrule    
    7 & 30 & 58630 & 47863 & 4941 & 688 & 34.99 & 9039 & 331 & 39.65 & 4098 & 357 \\
    \midrule    
    8 & 40 & 78488 & 65655 & 5943 & 732 & 44.33 & 13510 & 394 & 53.11 & 7567 & 338 \\
\bottomrule
    \end{tabular}
 }
\end{table}

Across all classes, the fair model achieves a notable increase in fairness except for the first class. For other classes, the average gain in fairness goes from 112 to 357. This comes with a cost, which increases significantly with the size of the instances, with up to 7567 additional changes on average for class 8. As instance size and complexity grow, the trade-offs between fairness and the cost to achieve it become more pronounced. This suggests that the more collaborative arcs, the higher the potential of achieving more fairness among schedulers. Overall, these results indicate that while fairness can be effectively integrated, its cost varies depending on problem structure, and some instances allow fair outcomes without compromising efficiency.

\subsection{Trade-off Between Efficiency and Fairness}

A weighted-objective model is also implemented, in which $\alpha$ represents the weight assigned to the efficient objective and $1 - \alpha$ corresponds to the weight assigned to the fair objective. Table \ref{Trade-off between Efficiency and Fairness} shows the results for various values of $\alpha$. The results show that as $\alpha$ increases from 0 to 1, the focus shifts from minimizing total changes ($\Delta$) to limiting the maximum disruption for any scheduler ($Z^*$). In Class 1, $\Delta$ increases from 321 at $\alpha=0$ to 357 at $\alpha=1$. In comparison, $Z^*$ decreases from 199 to 188 for intermediate values but does not improve under full fairness, indicating diminishing returns for fairness improvement. In Class 8, the total changes rise significantly from 5943 to 13510 as fairness becomes dominant, while $Z^*$ stabilizes at 394. This pattern highlights the cost of fairness: reducing the worst-case burden for individual schedulers requires spreading disruptions more widely, inflating the overall number of changes.

\begin{table}[t!]
  \centering
  \scalebox{0.59}{
  \caption{Trade-off between Efficiency and Fairness}
  \label{Trade-off between Efficiency and Fairness}
    \begin{tabular}{c|c|c|c|c|c|c|c|c|c|c|c|c|c|c|c|c|c|c}
    \toprule
    \multirow{2}[4]{*}{\textbf{Class}} & \multirow{2}[4]{*}{\textbf{\# Sched.}} & \multirow{2}[4]{*}{\textbf{I$_0$}} & \multirow{2}[4]{*}{\textbf{I$^*$}} & \multicolumn{3}{c|}{\textbf{Efficency ($\alpha=0$)}} & \multicolumn{3}{c|}{\textbf{$\alpha=0.25$}} & \multicolumn{3}{c|}{\textbf{$\alpha=0.5$}} & \multicolumn{3}{c|}{\textbf{$\alpha=0.75$}} & \multicolumn{3}{c}{\textbf{Fairness ($\alpha=1$)}} \\
\cmidrule{5-19}              &       &       &       & \textbf{$\Delta^*$} & \textbf{$Z^*$} & \textbf{Time} & \textbf{$\Delta^{0.25}$} & \textbf{$Z^{0.25}$} & \textbf{Time} & \textbf{$\Delta^{0.50}$} & \textbf{$Z^{0.50}$} & \textbf{Time} & \textbf{$\Delta^{0.75}$} & \textbf{$Z^{0.75}$} & \textbf{Time} & \textbf{$\Delta^{FAIR}$} & \textbf{$Z^{FAIR}$} & \textbf{Time} \\
    \midrule
    1 & 2 & 4236 & 3444 & 321 & 199 & 1.44 & 321 & 190 & 1.44 & 322 & 190 & 1.43 & 323 & 188 & 1.51 & 357 & 188 & 1.53 \\
    \midrule
    2 & 5 & 10644 & 8357 & 952 & 354 & 4.74 & 952 & 257 & 4.81 & 958 & 250 & 4.91 & 969 & 242 & 5.33 & 1135 & 242 & 5.50 \\
    \midrule
    3 & 8 & 18370 & 14709 & 1542 & 452 & 8.27 & 1542 & 303 & 9.30 & 1555 & 289 & 9.17 & 1574 & 276 & 9.28 & 1852 & 274 & 11.30 \\
    \midrule
    4 & 11 & 22733 & 17667 & 2048 & 443 & 12.25 & 2048 & 323 & 12.53 & 2057 & 314 & 13.03 & 2094 & 291 & 14.25 & 2796 & 285 & 15.50 \\
    \midrule
    5 & 15 & 28879 & 22863 & 2701 & 486 & 15.14 & 2701 & 360 & 15.98 & 2724 & 335 & 16.75 & 2754 & 311 & 17.04 & 4184 & 308 & 18.16 \\
    \midrule
    6 & 20 & 40799 & 32641 & 3725 & 710 & 22.13 & 3725 & 488 & 21.90 & 3764 & 443 & 21.85 & 3823 & 400 & 26.61 & 7350 & 397 & 24.98 \\
    \midrule
    7 & 30 & 58630 & 47863 & 4941 & 688 & 34.99 & 4941 & 428 & 33.89 & 4984 & 382 & 34.26 & 5041 & 341 & 37.58 & 9039 & 331 & 39.65 \\
    \midrule
    8 & 40 & 78488 & 65655 & 5943 & 732 & 44.33 & 5944 & 457 & 44.31 & 5970 & 425 & 47.86 & 6002 & 399 & 44.57 & 13510 & 394 & 53.11 \\
    \bottomrule
    \end{tabular}
    }
\end{table}

Execution time exhibits a clear dependency on both class size and fairness requirements. For Class 1, runtime remains relatively stable (less than 2 seconds) across all $\alpha$ values, implying that fairness enforcement is computationally inexpensive in small systems. However, in Class 8, where the number of schedulers and complexity increase, execution time grows as $\alpha$ approaches 1. This indicates that incorporating fairness constraints in large-scale scheduling not only raises the total number of changes but also increases computational effort. Overall, the table underscores that fairness reduces the burden on the most affected scheduler but at the expense of greater overall disruption and higher computational cost, especially in complex, multi-scheduler scenarios.

\begin{figure}[t!]
    \centering
    \begin{subfigure}[t]{0.48\textwidth}
        \centering
        \begin{tikzpicture}
        \begin{axis}[
            width=\textwidth,
            height=\textwidth,
            xlabel={\(\alpha\) (fairness weight)},
            ylabel={Percent Change vs Baseline (\(\alpha\)=0)},
            xmin=-0.05, xmax=1.05,
            ymin=-10, ymax=100,
            xtick={0.0, 0.2, 0.4, 0.6, 0.8, 1.0},
            ytick={0, 20, 40, 60, 80, 100},
            ymajorgrids=true,
            grid style=dashed,
            legend style={
                at={(0.02,0.98)},
                anchor=north west,
                draw=black,
                fill=white,
                font=\small
            }
        ]
        \addplot[color=blue, mark=*, thick] coordinates {
            (0.0, 0.0) (0.25, 0.0) (0.5, 0.0) (0.75, 1.0) (1.0, 88.8)
        };
        \addlegendentry{$\Delta$ Evolution (\(\Delta\)\% vs \(\alpha\)=0)}
        \addplot[color=orange, mark=square*, thick] coordinates {
            (0.0, 0.0) (0.25, 12.9) (0.5, 14.9) (0.75, 17.3) (1.0, 17.7)
        };
        \addlegendentry{Z Evolution (Z\% vs \(\alpha\)=0)}
        \end{axis}
        \end{tikzpicture}
        \caption{Trade-off Between Efficiency and Fairness}
        \label{fig:efficiency_fairness_tradeoff_2}
    \end{subfigure}
    \hfill
    \begin{subfigure}[t]{0.48\textwidth}
        \centering
        \begin{tikzpicture}
        \begin{axis}[
            xlabel={\(\alpha\) (fairness weight)},
            ylabel={Percent Share of Total Burden (\%)},
            width=\textwidth,
            height=\textwidth,
            xmin=0, xmax=1,
            ymin=0, ymax=100,
            xtick={0, 0.25, 0.5, 0.75, 1.0},
            ytick={0, 20, 40, 60, 80, 100},
            yticklabel={\pgfmathprintnumber{\tick}\%},
            stack plots=y,
            area style,
            enlarge x limits=false,
            axis background/.style={fill=gray!15},
        ]
        \addplot [fill=myblue, draw=none] coordinates {(0,5) (0.25,8) (0.5,6) (0.75,8) (1,12)} \closedcycle;
        \addplot [fill=myorange, draw=none] coordinates {(0,47) (0.25,40) (0.5,40) (0.75,35) (1,14)} \closedcycle;
        \addplot [fill=mylightgreen, draw=none] coordinates {(0,3) (0.25,5) (0.5,4) (0.75,5) (1,14)} \closedcycle;
        \addplot [fill=mypurple, draw=none] coordinates {(0,9) (0.25,9) (0.5,10) (0.75,10) (1,10)} \closedcycle;
        \addplot [fill=mytan, draw=none] coordinates {(0,4) (0.25,5) (0.5,6) (0.75,7) (1,13)} \closedcycle;
        \addplot [fill=mygray, draw=none] coordinates {(0,12) (0.25,12) (0.5,12) (0.75,12) (1,12)} \closedcycle;
        \addplot [fill=myyellow, draw=none] coordinates {(0,9) (0.25,9) (0.5,9) (0.75,10) (1,12)} \closedcycle;
        \addplot [fill=mycyan, draw=none] coordinates {(0,11) (0.25,12) (0.5,13) (0.75,13) (1,13)} \closedcycle;
        \end{axis}
        \end{tikzpicture}
        \caption{Burden Reallocation Across Schedulers}
        \label{fig:burden_reallocation}
    \end{subfigure}
    \caption{Side-by-side Comparison of Efficiency/Fairness Trade-off and Burden reallocation with $\alpha$}
    \label{fig:efficiency_fairness_tradeoff}
\end{figure}
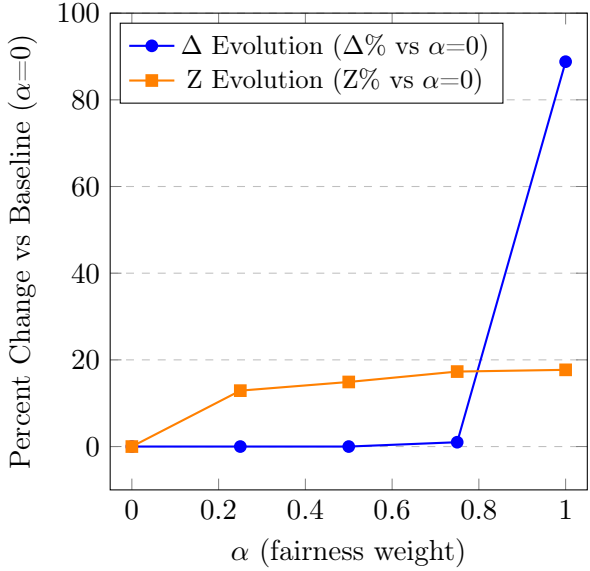
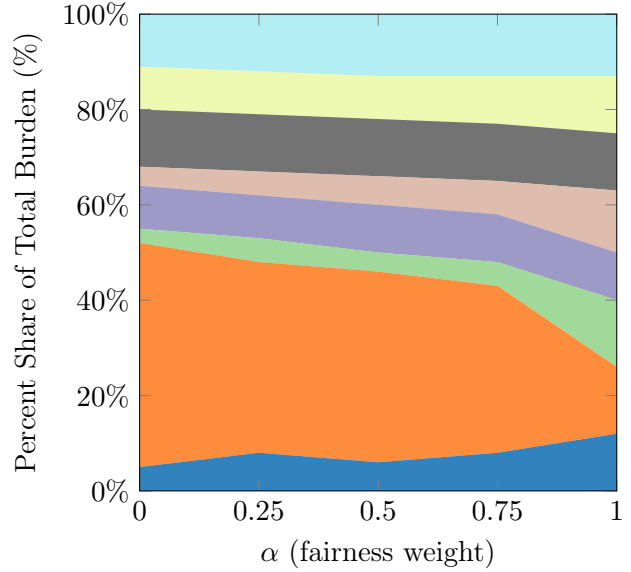

Figure \ref{fig:efficiency_fairness_tradeoff_2} displays two plots showing the percent change in $\Delta$ and $Z$ compared to the baseline ($\alpha=0$) for a representative instance as a function of the fairness level (various values of using $\alpha$). Figure \ref{fig:burden_reallocation} displays a stacked area chart showing the distribution of changes from Stage 2 across eight schedulers as a function of the fairness level (various values of using $\alpha$). The y-axis shows the percentage of total changes allocated to each scheduler. Each colored band corresponds to a specific scheduler. The width of each band at any fairness level reflects that scheduler’s share of the total changes at that level of fairness. The x-axis represents increasing fairness levels (from 0.0 to 1.0).

As fairness increases, the reduction in the number of maximum changes by schedulers increases to reach around 17.3\% while maintaining the same number of total changes. However, beyond a threshold ($\alpha=0.75$), the total number of changes increases significantly while ensuring a slight increase in the reduction in the number of maximum changes by schedulers. This shift reflects a redistribution of changes toward a more equitable allocation. Initially, scheduler 2 dominates with over 45\% of changes at the lowest fairness level, indicating high inequality in burden. However, its share declines significantly as fairness rises, while underutilized schedulers such as scheduler 1 and scheduler 3 gain a larger share. This shift reflects a redistribution of changes toward a more equitable allocation. The chart clearly shows how incorporating fairness into the objective leads to a more even distribution of scheduling changes among schedulers.

In addition to the Minimax metric, the analysis incorporates a fairness measure derived from the Gini coefficient \citep{fair_Guide_ChenHooker}. A detailed formulation and additional experimental results using this Gini metric are provided in Appendix \ref{B}.

\subsection{ML Impact}

The ML component has a significant impact on the computational performance presented above. Table \ref{tab:ML Impact Stage 1} illustrates the statistics for Stage 1, comparing scenarios with and without ML, and using the top-5 equipment suggestions provided by the ML model for each arc. Without ML, most instances in Classes 30 and 40 could not be solved. Furthermore, by learning from historical data, ML reduces the number of arcs by approximately 83\%, which enables the solver to prove optimality on the reduced problem in cases where this was not possible otherwise and leads to a 94\% reduction in execution time. Similarly to Stage 1, ML reduces the number of arcs considered by the Stage 2 models by approximately 75\%, leading to a substantial 90\% reduction in execution time, and improves the solver’s ability to close optimality gaps. The ML component, therefore, does not alter the optimal solution of the original OR model; instead, it identifies a smaller and more relevant decision space that allows the solver to reach optimality more consistently and more efficiently on the reduced problem.

\begin{table}[!t]
  \centering
  \scalebox{0.75}{
  \caption{ML Impact Stage 1}
  \label{tab:ML Impact Stage 1}
    \begin{tabular}{c|c|c|c|c|c|c|c|c|c}
    \toprule
    \multirow{2}[4]{*}{\textbf{Class}} & \multirow{2}[4]{*}{\textbf{\# Sched.}} & \multirow{2}[4]{*}{\textbf{I$_0$}} & \multirow{2}[4]{*}{\textbf{I$^*$}} & \multicolumn{2}{c|}{\textbf{Without ML}} & \multicolumn{2}{c|}{\textbf{With ML}} & \multirow{2}[4]{*}{\textbf{GainA}} & \multirow{2}[4]{*}{\textbf{GainT}} \\
    \cline{5-8}
      &       &       &       & \textbf{Arcs} & \textbf{Time} & \textbf{Arcs} & \textbf{Time} &       &  \\
    \midrule
    1 & 2 & 4236 & 3444 & 138992 & 8.94 & 48971 & 1.02 & 64.77\% & 75.17\% \\
    \midrule 
    2 & 5 & 10644 & 8357 & 462500 & 58.70 & 127236 & 3.13 & 75.03\% & 91.12\% \\
    \midrule 
    3 & 8 & 18370 & 14709 & 893028 & 137.19 & 206595 & 5.53 & 76.87\% & 92.24\% \\
    \midrule    
    4 & 11 & 22733 & 17667 & 1278907 & 277.74 & 278858 & 7.61 & 78.20\% & 94.77\% \\
    \midrule    
    5 & 15 & 28879 & 22863 & 1816934 & 866.41 & 189891 & 9.57 & 89.55\% & 98.90\%  \\
    \midrule    
    6 & 20 & 40799 & 32641 & 3004375 & 3468.60 & 270678 & 13.19 & 90.99\% & 98.93\% \\
    \midrule    
    7 & 30 & 58630 & 47863 & 5259231 & - & 401788 & 19.77 & 92.36\% & 98.96\% \\
    \midrule    
    8 & 40 & 78488 & 65655 & 8341357 & - & 562008 & 26.48 & 93.26\% & 98.97\% \\
\bottomrule
    \end{tabular}
 }
\end{table}

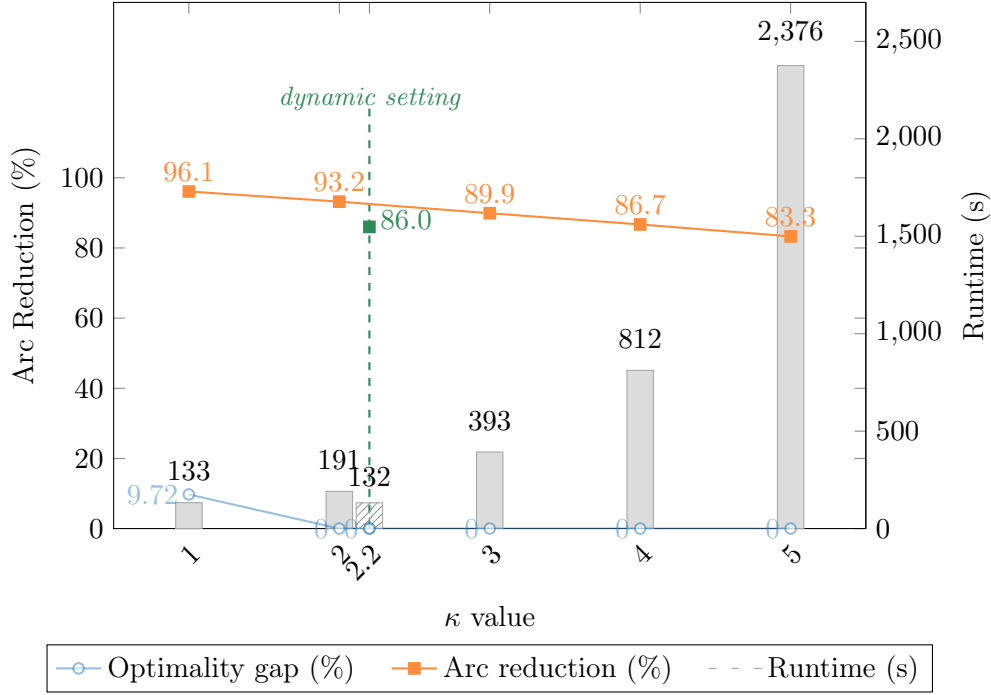
\begin{figure}[t!]
\centering
\begin{tikzpicture}
\begin{axis}[
  name=background,
  width=10cm, height=7cm,
  scale only axis,
  axis line style={on layer=axis foreground},
  axis y line*=right,
  axis x line=none,
  ylabel={Runtime (s)},
  xmin=0.5, xmax=5.5,
  ymin=0, ymax=2700,
  xtick=\empty,
  clip=false,
]
  \addplot[
    ybar, fill=lightgray, draw=gray!70,
    nodes near coords,
    nodes near coords align={center},
    every node near coord/.style={
        font=\normalsize,
        yshift=12pt,             
        on layer=axis foreground
    }
  ] coordinates {(1,133) (2,191) (3,393) (4,812) (5,2376)};
  \addplot[
    ybar, draw=gray!70, pattern=north east lines, pattern color=gray!80,
    nodes near coords,
    nodes near coords align={center},
    every node near coord/.style={
        font=\normalsize,
        yshift=10pt,             
        on layer=axis foreground
    }
  ] coordinates {(2.2,132)};
\end{axis}
\begin{axis}[
  at={(background.south west)},
  anchor=south west,
  width=10cm, height=7cm,
  scale only axis,
  xlabel={$\kappa$ value},
  ylabel={Arc Reduction (\%)},
  xmin=0.5, xmax=5.5,
  ymin=0, ymax=150,
  xtick={1,2,2.2,3,4,5},
  xticklabel style={rotate=45, anchor=north east},
  ytick={0,20,40,60,80,100},
  axis lines=box,
  legend style={
    at={(0.5,-0.22)}, 
    anchor=north,
    legend columns=-1,
    /tikz/every even column/.append style={column sep=0.4cm}
},
  set layers,
]
  \begin{pgfonlayer}{axis grid}
    \draw[mygreen, dashed, thick] (axis cs:2.2,0) -- (axis cs:2.2,120);
  \end{pgfonlayer}
  \node[color=mygreen, font=\small\itshape, anchor=south, yshift=2pt]
    at (axis cs:2.2,115) {dynamic setting};
  \addplot[
    color=myblue, mark=*, thick,
    opacity=0.5,                
    mark options={draw=myblue, fill=white, opacity=0.5}, 
    nodes near coords,
    nodes near coords align={left},
  ] coordinates
    {(1,9.72) (2,0.00) (2.2,0.00) (3,0.00) (4,0.00) (5,0.00)};
  \addlegendentry{Optimality gap (\%)}
  \addplot[
    color=myorange, mark=square*, thick,
    nodes near coords,
    nodes near coords align={above},
  ] coordinates
    {(1,96.10) (2,93.20) (3,89.9) (4,86.70) (5,83.30)};
  \addlegendentry{Arc reduction (\%)}
  \addlegendimage{ybar, fill=lightgray, draw=gray!70}
  \addlegendentry{Runtime (s)}
  \addplot[
    only marks, mark=*, mark options={fill=white, draw=myblue, thick},
    opacity=0.5, 
    forget plot
  ] coordinates {(2.2,0.00)};
\addplot[
    only marks,
    mark=square*,
    mark options={fill=mygreen, draw=mygreen, thick},
    very thick,
    forget plot,
] coordinates {(2.2, 86.0)}
    node[font=\normalsize, anchor=south west, yshift=-5pt, text=mygreen] {86.0};
\end{axis}
\end{tikzpicture}
\caption{Dynamic-$\kappa$ versus Static-$\kappa$}
\label{fig:kappa}
\end{figure}

To evaluate the benefits of the proposed dynamic top-\(\kappa_a\) mechanism, the comparison is made with a traditional static top-\(\kappa\) approach, in which a fixed \(\kappa\) value is applied uniformly across all arcs in a full network comprising 50 schedulers, 8000 nodes, and 24 million arcs. Figure~\ref{fig:kappa} illustrates how optimality gap, arc reduction, and runtime execution vary with static \(\kappa \in \{1,5\}\). The analysis highlights the trade-offs between solution quality and computational efficiency: smaller \(\kappa\) values yield faster but suboptimal solutions (e.g., 9.72\% optimality gap for $\kappa=1$), whereas larger \(\kappa\) values achieve optimal solutions at the expense of increased computation time. On the full network, when \(\kappa = 5\), the solution is optimal, but the solver takes 2376 seconds.

The dynamic \(\kappa_a\) results, shown in green in Figure~\ref{fig:kappa}, correspond to the adaptive mechanism in which each arc is assigned its own \(\kappa_a\) based on the betweenness metric. The lone green point in the figure, with $\kappa$ set to 2.2, is the weighted average of \(\kappa_a\) in the full network instance. Under this dynamic rule, the runtime is approximately 132 seconds, the arc reduction is 86.00\%, and the solution remains optimal.

Results highlight how adapting $\kappa_a$ to arc betweenness centrality affects the scalability of the OR models and the quality of the resulting substitution plans. Indeed, this adaptive mechanism tailors the search space to the local network structure, enhancing scalability and computational efficiency without compromising solution quality. By dynamically assigning values of $\kappa$ to arcs based on the betweenness degree, this allows larger-scale networks to retain optimality while achieving runtimes consistent with smaller instances that rely on a static $\kappa$.

\subsection{Managerial Insights}

The results highlight a fundamental trade-off in resource substitution: optimizing solely for efficiency (e.g., minimizing total changes) can lead to unequal distribution of workload among schedulers. While such solutions are efficient from an optimization perspective, they may disproportionately burden certain schedulers, which in practice can result in partial implementation of recommendations, as some changes exceed what schedulers are willing or able to accommodate. Conversely, incorporating fairness into the objective promotes a more equitable distribution of tasks, often with only a modest increase in cost or computational effort. This substantially reduces disparities among schedulers, demonstrating that fairness-aware optimization can produce solutions that remain operationally sound while greatly improving workforce satisfaction and sustainability.

Importantly, incorporating fairness transforms the solution space from a single optimal point into a rich portfolio of efficient-fair solutions. Weighted-objective models allow decision-makers to balance efficiency and fairness, generating a Pareto frontier from which schedulers can select the most suitable solution for their operational context. For example, a company may accept a slight increase in total cost or complexity to achieve a fairer distribution of workload, particularly during peak periods or volatile conditions. This approach supports human-centered decision-making, where long-term resilience and employee satisfaction outweigh marginal efficiency gains. 

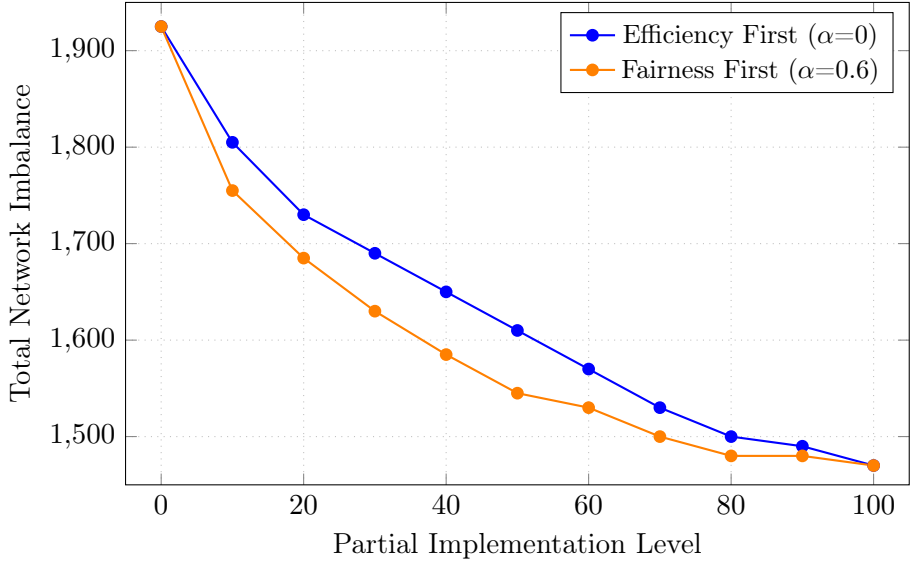
\begin{figure}[ht!]
\centering
\begin{tikzpicture}
\begin{axis}[
    xlabel={Partial Implementation Level},
    ylabel={Total Network Imbalance},
    xmin=-5, xmax=105,
    ymin=1450, ymax=1950,
    xtick={0,20,40,60,80,100},
    ytick={1500,1600,1700,1800,1900},
    grid=major,
    grid style={dotted},
    width=12cm,
    height=8cm,
    legend style={
        at={(0.98,0.98)},       
        anchor=north east,      
        draw=black,             
        fill=white,             
        font=\small,            
    },
]
\addplot[
    color=blue,
    mark=*,
    thick,
    ]
    coordinates {
    (0, 1925)
    (10, 1805)
    (20, 1730)
    (30, 1690)
    (40, 1650)
    (50, 1610)
    (60, 1570)
    (70, 1530)
    (80, 1500)
    (90, 1490)
    (100, 1470)
    };
\addlegendentry{Efficiency First ($\alpha$=0)}
\addplot[
    color=orange,
    mark=*,
    thick,
    ]
    coordinates {
    (0, 1925)
    (10, 1755)
    (20, 1685)
    (30, 1630)
    (40, 1585)
    (50, 1545)
    (60, 1530)
    (70, 1500)
    (80, 1480)
    (90, 1480)
    (100, 1470)
    };
\addlegendentry{Fairness First ($\alpha$=0.6)}
\end{axis}
\end{tikzpicture}
\caption{Comparison between the Pareto Frontiers for Efficiency and Fairness under a Partial Implementation Strategy}
\label{fig:partial implementation}
\end{figure}

The Pareto frontier of a representative instance is visualized in Figure~\ref{fig:partial implementation}. Adopting a 60\% minimax fairness objective (\(\alpha = 0.6\)) results in a substantial rebalancing of workload without any loss in global solution quality. Specifically, the burden on the most-impacted scheduler decreases by 42\% from 732 to 425. This demonstrates that substitution can be effectively redistributed to less-burdened neighbors without a trade-off.

In practice, the friction of coordinating among schedulers means that only a fraction of recommended changes are ever implemented, a setting referred to here as \emph{partial implementation}, where schedulers adopt only the top-priority substitutions rather than the full set. As simulated in Figure~\ref{fig:partial implementation}, this is where the fair solution excels. When prioritizing and implementing only the highest-impact substitutions, the fair solution yields a greater reduction in system imbalance. This occurs because the fair algorithm had already allocated high-value changes more evenly across all schedulers, making fair partial implementation a more effective strategy.

Another important aspect to highlight is that the degree of collaboration among schedulers, defined as the extent to which schedulers share resources and coordinate substitutions across different regions, has a pronounced impact on both computational performance and fairness outcomes. When schedulers collaborate extensively across regions, the resulting network exhibits strong inter-regional connectivity, enabling flexible resource sharing and more balanced workload distributions. In contrast, limited collaboration results in a fragmented structure with few inter-regional exchanges, thereby constraining adaptability and reinforcing local imbalances. As collaboration increases, the model can redistribute substitutions more evenly, leading to fairer and more stable solutions across the network. These findings underscore that collaborative flexibility is a key driver of fairness and efficiency in large-scale resource allocation systems.

\section{Conclusion}\label{section:5}

Ensuring that the right resource is available at the right location and time remains a critical challenge in large-scale logistics networks, largely due to persistent imbalances arising from fluctuating demand and dynamic product flows. This paper addresses the NP-hard resource substitution problem by proposing a generic hybrid framework that integrates operations research and machine learning. The framework explicitly models operational constraints, incorporates fairness considerations to align centralized plans with decentralized scheduler preferences, and leverages historical data to guide and accelerate the search for high-quality solutions. Applied to the network of one of the largest package delivery companies in the world, the framework demonstrated superior performance relative to existing substitution-based methods, achieving both computational efficiency and practical relevance. Beyond package delivery, the approach is broadly applicable to a wide range of resource balancing problems in transportation and logistics networks, including container repositioning, fleet assignment, and bike-sharing rebalancing. Future research directions include incorporating stochastic demand and supply uncertainties, exploring adaptive online learning approaches, and developing distributed implementations to further enhance scalability and operational flexibility in large-scale, real-time settings.

\section*{Acknowledgments}

This research was partly supported by the NSF AI Institute for Advances in Optimization (Award 2112533).

\begin{spacing}{1}
\typeout{}
\bibliographystyle{apalike}
\bibliography{References.bib}

@article{yang2022substitution,
  title={Substitution-based equipment balancing in service networks with multiple equipment types},
  author={Yang, Yu and Ridouane, Yassine and Boland, Natashia and Erera, Alan and Savelsbergh, Martin},
  journal={European Journal of Operational Research},
  volume={300},
  number={3},
  pages={966--978},
  year={2022},
  publisher={Elsevier}
}

@inproceedings{leddon1967scheduling,
  title={Scheduling empty freight car fleets on the {L}ouisville and {N}ashville railroad},
  author={Leddon, CD and Wrathall, E},
  booktitle={Second International Symposium on the Use of Cybernetics on the Railways},
  pages={1--6},
  year={1967}
}

@article{misra1972linear,
  title={Linear programming of empty wagon disposition},
  author={Misra, SC},
  journal={Rail International},
  volume={3},
  number={3},
  year={1972}
}

@article{white1972dynamic,
  title={{D}ynamic transshipment networks: {A}n algorithm and its application to the distribution of empty containers},
  author={White, William W},
  journal={Networks},
  volume={2},
  number={3},
  pages={211--236},
  year={1972},
  publisher={Wiley Online Library}
}

@article{abrache1999new,
  title={New decomposition algorithm for the deterministic dynamic allocation of empty containers},
  author={Abrache, Jawad and Crainic, Teodor G and Gendreau, Michel},
  year={1999}
}

@article{erera2005global,
  title={Global intermodal tank container management for the chemical industry},
  author={Erera, Alan L and Morales, Juan C and Savelsbergh, Martin},
  journal={Transportation Research Part E: Logistics and Transportation Review},
  volume={41},
  number={6},
  pages={551--566},
  year={2005},
  publisher={Elsevier}
}

@article{powell1986stochastic,
  title={A stochastic model of the dynamic vehicle allocation problem},
  author={Powell, Warren B},
  journal={Transportation science},
  volume={20},
  number={2},
  pages={117--129},
  year={1986},
  publisher={INFORMS}
}

@article{crainic1993dynamic,
  title={Dynamic and stochastic models for the allocation of empty containers},
  author={Crainic, Teodor Gabriel and Gendreau, Michel and Dejax, Pierre},
  journal={Operations research},
  volume={41},
  number={1},
  pages={102--126},
  year={1993},
  publisher={INFORMS}
}

@article{powell1987operational,
  title={An operational planning model for the dynamic vehicle allocation problem with uncertain demands},
  author={Powell, Warren B},
  journal={Transportation Research Part B: Methodological},
  volume={21},
  number={3},
  pages={217--232},
  year={1987},
  publisher={Elsevier}
}

@article{nair2011fleet,
  title={Fleet management for vehicle sharing operations},
  author={Nair, Rahul and Miller-Hooks, Elise},
  journal={Transportation Science},
  volume={45},
  number={4},
  pages={524--540},
  year={2011},
  publisher={INFORMS}
}

@article{shu2013models,
  title={Models for effective deployment and redistribution of bicycles within public bicycle-sharing systems},
  author={Shu, Jia and Chou, Mabel C and Liu, Qizhang and Teo, Chung-Piaw and Wang, I-Lin},
  journal={Operations Research},
  volume={61},
  number={6},
  pages={1346--1359},
  year={2013},
  publisher={INFORMS}
}

@article{bruglieri2014vehicle,
  title={The vehicle relocation problem for the one-way electric vehicle sharing: an application to the {M}ilan case},
  author={Bruglieri, Maurizio and Colorni, Alberto and Lue, Alessandro},
  journal={Procedia-Social and Behavioral Sciences},
  volume={111},
  pages={18--27},
  year={2014},
  publisher={Elsevier}
}

@article{ghosh2017dynamic,
  title={Dynamic repositioning to reduce lost demand in bike sharing systems},
  author={Ghosh, Supriyo and Varakantham, Pradeep and Adulyasak, Yossiri and Jaillet, Patrick},
  journal={Journal of Artificial Intelligence Research},
  volume={58},
  pages={387--430},
  year={2017}
}

@inproceedings{li2018dynamic,
  title={Dynamic bike reposition: A spatio-temporal reinforcement learning approach},
  author={Li, Yexin and Zheng, Yu and Yang, Qiang},
  booktitle={Proceedings of the 24th ACM SIGKDD International Conference on Knowledge Discovery \& Data Mining},
  pages={1724--1733},
  year={2018}
}

@article{lv2020hybrid,
  title={A hybrid algorithm for the static bike-sharing re-positioning problem based on an effective clustering strategy},
  author={Lv, Chang and Zhang, Chaoyong and Lian, Kunlei and Ren, Yaping and Meng, Leilei},
  journal={Transportation Research Part B: Methodological},
  volume={140},
  pages={1--21},
  year={2020},
  publisher={Elsevier}
}

@inproceedings{chung2018bike,
  title={Bike angels: An analysis of {Citi Bike}'s incentive program},
  author={Chung, Hangil and Freund, Daniel and Shmoys, David B},
  booktitle={Proceedings of the 1st ACM SIGCAS Conference on Computing and Sustainable Societies},
  pages={1--9},
  year={2018}
}

@article{schuijbroek2017inventory,
  title={Inventory rebalancing and vehicle routing in bike sharing systems},
  author={Schuijbroek, Jasper and Hampshire, Robert C and Van Hoeve, W-J},
  journal={European Journal of Operational Research},
  volume={257},
  number={3},
  pages={992--1004},
  year={2017},
  publisher={Elsevier}
}

@article{du1997fleet,
  title={Fleet sizing and empty equipment redistribution for center-terminal transportation networks},
  author={Du, Yafeng and Hall, Randolph},
  journal={Management Science},
  volume={43},
  number={2},
  pages={145--157},
  year={1997},
  publisher={INFORMS}
}

@article{song2005optimal,
  title={Optimal threshold control of empty vehicle redistribution in two depot service systems},
  author={Song, Dong-Ping},
  journal={IEEE Transactions on Automatic Control},
  volume={50},
  number={1},
  pages={87--90},
  year={2005},
  publisher={IEEE}
}

@article{jansen2004operational,
  title={Operational planning of a large-scale multi-modal transportation system},
  author={Jansen, Benjamin and Swinkels, Pieter CJ and Teeuwen, Geert JA and de Fluiter, Babette van Antwerpen and Fleuren, Hein A},
  journal={European Journal of Operational Research},
  volume={156},
  number={1},
  pages={41--53},
  year={2004},
  publisher={Elsevier}
}

@article{erera2009robust,
  title={Robust optimization for empty repositioning problems},
  author={Erera, Alan L and Morales, Juan C and Savelsbergh, Martin},
  journal={Operations Research},
  volume={57},
  number={2},
  pages={468--483},
  year={2009},
  publisher={INFORMS}
}

@article{long2012sample,
  title={The sample average approximation method for empty container repositioning with uncertainties},
  author={Long, Yin and Lee, Loo Hay and Chew, Ek Peng},
  journal={European Journal of Operational Research},
  volume={222},
  number={1},
  pages={65--75},
  year={2012},
  publisher={Elsevier}
}

@article{xie2017empty,
  title={Empty container management and coordination in intermodal transport},
  author={Xie, Yangyang and Liang, Xiaoying and Ma, Lijun and Yan, Houmin},
  journal={European Journal of Operational Research},
  volume={257},
  number={1},
  pages={223--232},
  year={2017},
  publisher={Elsevier}
}

@incollection{lourencco2003iterated,
  title={Iterated local search},
  author={Louren{\c{c}}o, Helena R and Martin, Olivier C and St{\"u}tzle, Thomas},
  booktitle={Handbook of metaheuristics},
  pages={320--353},
  year={2003},
  publisher={Springer}
}

@book{desaulniers2006column,
  title={Column generation},
  author={Desaulniers, Guy and Desrosiers, Jacques and Solomon, Marius M},
  volume={5},
  year={2006},
  publisher={Springer Science \& Business Media}
}

@article{lopez2016irace,
  title={The irace package: {I}terated racing for automatic algorithm configuration},
  author={L{\'o}pez-Ib{\'a}{\~n}ez, Manuel and Dubois-Lacoste, J{\'e}r{\'e}mie and C{\'a}ceres, Leslie P{\'e}rez and Birattari, Mauro and St{\"u}tzle, Thomas},
  journal={Operations Research Perspectives},
  volume={3},
  pages={43--58},
  year={2016},
  publisher={Elsevier}
}

@article{himmich2023mpils,
  title={{MPILS}: {A}n automatic tuner for {MILP} solvers},
  author={Himmich, Ilyas and El Hachemi, Nizar and El Hallaoui, Issma{\"\i}l and Metrane, Abdelmoutalib and Soumis, Fran{\c{c}}ois and others},
  journal={Computers \& Operations Research},
  volume={159},
  pages={106344},
  year={2023},
  publisher={Elsevier}
}

@book{goodfellow2016deep,
  title={Deep learning},
  author={Goodfellow, Ian and Bengio, Yoshua and Courville, Aaron},
  year={2016},
  publisher={MIT press}
}

@article{lecun2015deep,
  title={Deep learning},
  author={LeCun, Yann and Bengio, Yoshua and Hinton, Geoffrey},
  journal={nature},
  volume={521},
  number={7553},
  pages={436--444},
  year={2015},
  publisher={Nature Publishing Group UK London}
}

@article{intro_businessreport_atri,
  title={An Analysis of the Operational Costs of Trucking: 2024 Update},
  author={{ATRI}},
  journal={ATRI Research Report},
  year={2024},
  note={Available at \url{https://truckingresearch.org/wp-content/uploads/2024/06/ATRI-Operational-Cost-of-Trucking-06-2024.pdf}}
}

@article{intro_trucking_leslie,
  title={An Analysis of the Operational Costs of Trucking: 2024 Update},
  author={Leslie, Alex and Murray, Dan},
  year={2024}
}

@article{intro_airline_jarrah,
  title={An efficient airline re-fleeting model for the incremental modification of planned fleet assignments},
  author={Jarrah, Ahmad I and Goodstein, Jon and Narasimhan, Ram},
  journal={Transportation Science},
  volume={34},
  number={4},
  pages={349--363},
  year={2000},
  publisher={INFORMS}
}

@article{intro_shipping_lin,
  title={The container retrieval problem with respect to relocation},
  author={Lin, Dung-Ying and Lee, Yen-Ju and Lee, Yusin},
  journal={Transportation Research Part C: Emerging Technologies},
  volume={52},
  pages={132--143},
  year={2015},
  publisher={Elsevier}
}

@article{intro_bikeBenchmark_dell,
  title={The bike sharing rebalancing problem: Mathematical formulations and benchmark instances},
  author={Dell'Amico, Mauro and Hadjicostantinou, Eleni and Iori, Manuel and Novellani, Stefano},
  journal={Omega},
  volume={45},
  pages={7--19},
  year={2014},
  publisher={Elsevier}
}

@article{fair_2agents_yu,
  title={Analyzing the price of fairness in scheduling problems with two agents},
  author={Yu, Jin and Liu, Peihai and Lu, Xiwen and Gu, Manzhan},
  journal={European Journal of Operational Research},
  volume={321},
  number={3},
  pages={750--759},
  year={2025},
  publisher={Elsevier}
}

@article{fair_PriceOfFairness_bertsimas,
  title={The price of fairness},
  author={Bertsimas, Dimitris and Farias, Vivek F. and Trichakis, Nikolaos},
  journal={Operations Research},
  volume={59},
  number={1},
  pages={17--31},
  year={2011},
  publisher={INFORMS}
}

@article{fair_InequityAverse_karsu,
  title={Inequity averse optimization in operational research},
  author={Karsu, {\"O}zlem and Morton, Alec},
  journal={European Journal of Operational Research},
  volume={245},
  number={2},
  pages={343--359},
  year={2015},
  publisher={Elsevier}
}

@article{fair_SocialWelfare_ChenHooker,
  title={Fairness through Social Welfare Optimization},
  author={Chen, Violet Xinying and Hooker, J. N.},
  journal={arXiv preprint arXiv:2102.00311},
  year={2022},
  publisher={arXiv},
  url={https://arxiv.org/abs/2102.00311}
}

@article{fair_Guide_ChenHooker,
  title={A guide to formulating fairness in an optimization model},
  author={Chen, Violet Xinying and Hooker, J. N.},
  journal={Annals of Operations Research},
  volume={326},
  number={1},
  pages={581--619},
  year={2023},
  publisher={Springer}
}

@article{fair_FacilityLocation_gupta,
  title={Which {LP} norm is the fairest? Approximations for fair facility location across all ``p''},
  author={Gupta, Swati and Moondra, Jai and Singh, Mohit},
  journal={Proceedings of the 24th ACM Conference on Economics and Computation},
  pages={817},
  year={2023},
  publisher={Association for Computing Machinery}
}

@article{ml_survey4opt_bengio,
  title={Machine learning for combinatorial optimization: A methodological tour d’horizon},
  author={Bengio, Yoshua and Lodi, Andrea and Prouvost, Antoine},
  journal={European Journal of Operational Research},
  volume={290},
  number={2},
  pages={405--421},
  year={2021},
  publisher={Elsevier}
}

@article{ml_column_morabit,
  title={Machine-Learning--Based Column Selection for Column Generation},
  author={Morabit, Mouad and Zarpellon, Giulia and Lodi, Andrea and Bengio, Yoshua and Gaudreault, Jean and Rousseau, Louis-Martin},
  journal={Transportation Science},
  volume={55},
  number={4},
  pages={815--831},
  year={2021},
  publisher={INFORMS}
}

@article{ml_arcselection_morabit,
  title={Machine-Learning--Based Arc Selection for Constrained Shortest Path Problems in Column Generation},
  author={Morabit, Mouad and Desaulniers, Guy and Lodi, Andrea},
  journal={INFORMS Journal on Optimization},
  volume={5},
  number={2},
  pages={191--210},
  year={2023},
  publisher={INFORMS}
}

@article{ml_aircrew_tahir,
  title={An Improved Integral Column Generation Algorithm Using Machine Learning for Aircrew Pairing},
  author={Tahir, Adil and Quesnel, Frédéric and Desaulniers, Guy and El Hallaoui, Issmail and Yaakoubi, Yassine},
  journal={Transportation Science},
  volume={55},
  number={6},
  pages={1411--1429},
  year={2021},
  publisher={INFORMS}
}

@article{freund2022minimizing,
  title={Minimizing multimodular functions and allocating capacity in bike-sharing systems},
  author={Freund, Daniel and Henderson, Shane G and Shmoys, David B},
  journal={Operations Research},
  volume={70},
  number={5},
  pages={2715--2731},
  year={2022},
  publisher={INFORMS}
}

@inproceedings{kingma2014adam,
  author    = {Diederik P. Kingma and
               Jimmy Ba},
  title     = {Adam: {A} Method for Stochastic Optimization},
  booktitle = {3rd International Conference on Learning Representations, {ICLR} 2015,
               San Diego, CA, USA, May 7-9, 2015, Conference Track Proceedings},
  year      = {2015},
  url       = {http://arxiv.org/abs/1412.6980},
  bibsource = {dblp computer science bibliography, https://dblp.org}
}

@article{akhlaghi2025propel,
  title={Propel: Supervised and reinforcement learning for large-scale supply chain planning},
  author={Akhlaghi, Vahid Eghbal and Zandehshahvar, Reza and Van Hentenryck, Pascal},
  journal={arXiv preprint arXiv:2504.07383},
  year={2025}
}

@article{wu2024towards,
  title={Towards resilience: Primal large-scale re-optimization},
  author={Er Raqabi, El Mehdi and Wu, Yong and El Hallaoui, Issma{\"\i}l and Soumis, Fran{\c{c}}ois and others},
  journal={Transportation Research Part E: Logistics and Transportation Review},
  volume={192},
  pages={103819},
  year={2024},
  publisher={Elsevier}
}

@article{scavuzzo2024machine,
  title={Machine learning augmented branch and bound for mixed integer linear programming},
  author={Scavuzzo, Lara and Aardal, Karen and Lodi, Andrea and Yorke-Smith, Neil},
  journal={Mathematical Programming},
  pages={1--44},
  year={2024},
  publisher={Springer}
}

@article{lodi2024framework,
  title={A framework for fair decision-making over time with time-invariant utilities},
  author={Lodi, Andrea and Sankaranarayanan, Sriram and Wang, Guanyi},
  journal={European Journal of Operational Research},
  volume={319},
  number={2},
  pages={456--467},
  year={2024},
  publisher={Elsevier}
}

@article{erraqabi1,
title = {Incremental {LNS} Framework for Integrated Production, Inventory, and Vessel Scheduling: Application to a Global Supply Chain},
author = {Er Raqabi, El Mehdi and Himmich, Ilyas and El Hachemi, Nizar and El Hallaoui, Issmaïl and Soumis, François},
journal = {Omega},
volume = {116},
pages = {102821},
year = {2023},
issn = {0305-0483},
doi = {https://doi.org/10.1016/j.omega.2022.102821},
url = {https://www.sciencedirect.com/science/article/pii/S0305048322002274},
}

@article{erraqabi5,
  title={{OCP} optimizes its supply chain for {A}frica},
  author={Er Raqabi, El Mehdi and Beljadid, Ahmed and Bennouna, Mohammed Ali and Bennouna, Rania and Boussaadi, Latifa and El Hachemi, Nizar and El Hallaoui, Issmail and Fender, Michel and Jamali, Mohamed Anouar and Si Hammou, Nabil and others},
  journal={INFORMS Journal on Applied Analytics},
  volume={55},
  number={6},
  pages={437--456},
  doi = {https://doi.org/10.1287/inte.2023.0073},
  year={2025},
  publisher={INFORMS}
}

@article{erraqabi2,
  title = {The {P}rimal {B}enders {D}ecomposition},
  author = {Er Raqabi, El Mehdi and El Hallaoui, Issmail and Soumis, François},
  pages = {},
  year = {2023},
  month = aug,
  month_numeric = {08},
  journal = {Les Cahiers du GERAD},
  volume = {G-2023-27},
  institution = {Groupe d’études et de recherche en analyse des décisions},
  address = {GERAD, Montréal QC H3T 2A7, Canada},
  URL = {https://www.gerad.ca/fr/papers/G-2023-27}
}

@book{newman2018networks,
  title={Networks},
  author={Newman, Mark},
  year={2018},
  publisher={Oxford University Press}
}

@article{quesnel2022deep,
  title={Deep-learning-based partial pricing in a branch-and-price algorithm for personalized crew rostering},
  author={Quesnel, Fr{\'e}d{\'e}ric and Wu, Alice and Desaulniers, Guy and Soumis, Fran{\c{c}}ois},
  journal={Computers \& Operations Research},
  volume={138},
  pages={105554},
  year={2022},
  publisher={Elsevier}
}
\end{spacing}    

\vspace{-0.3cm}

\begin{appendices}
\renewcommand{\thesection}{\Alph{section}}
\section{Compatibility Matrix Package Delivery Company} 
\label{A}

The company employs different types of equipment in its ground transportation network, both trailers and containers, which are grouped into categories based on their size: shorts (trailers with a length of 28 feet), longs (trailers with a length ranging from 40 to 48 feet), and extra longs (trailers with a length of 53 feet). 

\begin{table}[ht!]
\centering
\scalebox{0.9}{
\begin{tabular}{|c|cccccccccccccc|}
\hline
& $r_1$ & $r_2$ & $r_3$ & $r_4$ & $r_5$ & $r_6$ & $r_7$ & $r_8$ & $r_9$ & $r_{10}$ & $r_{11}$ & $r_{12}$ & $r_{13}$ & $r_{14}$ \\ \hline
$r_1$   & 1 & 1 & 1 & 1 & 0 & 0 & 0 & 0 & 0 & 0 & 0 & 0 & 0 & 0 \\
$r_2$  & 1 & 1 & 1 & 1 & 0 & 0 & 0 & 0 & 0 & 0 & 0 & 0 & 0 & 0 \\
$r_3$  & 1 & 1 & 1 & 1 & 0 & 0 & 0 & 0 & 0 & 0 & 0 & 0 & 0 & 0 \\
$r_4$  & 1 & 1 & 1 & 1 & 1 & 0 & 0 & 0 & 0 & 0 & 0 & 0 & 0 & 1 \\
$r_5$  & 1 & 1 & 1 & 1 & 1 & 0 & 0 & 0 & 0 & 0 & 0 & 0 & 0 & 1 \\
$r_6$   & 1 & 1 & 1 & 1 & 1 & 1 & 0 & 0 & 0 & 0 & 0 & 0 & 0 & 1 \\
$r_7$   & 1 & 1 & 1 & 1 & 1 & 1 & 1 & 0 & 0 & 0 & 0 & 0 & 0 & 1 \\
$r_8$  & 1 & 1 & 1 & 1 & 1 & 1 & 1 & 1 & 0 & 0 & 0 & 0 & 0 & 1 \\
$r_9$   & 1 & 1 & 1 & 1 & 1 & 1 & 1 & 1 & 1 & 0 & 0 & 0 & 0 & 1 \\
$r_{10}$  & 1 & 1 & 1 & 1 & 1 & 1 & 1 & 1 & 1 & 1 & 0 & 0 & 0 & 1 \\
$r_{11}$ & 1 & 1 & 1 & 1 & 1 & 1 & 1 & 1 & 1 & 1 & 1 & 0 & 0 & 1 \\
$r_{12}$ & 0 & 0 & 0 & 0 & 0 & 0 & 0 & 0 & 0 & 0 & 0 & 1 & 0 & 0 \\
$r_{13}$ & 1 & 1 & 1 & 1 & 1 & 1 & 1 & 1 & 1 & 1 & 1 & 0 & 1 & 1 \\
$r_{14}$  & 1 & 1 & 1 & 1 & 1 & 1 & 1 & 1 & 1 & 1 & 1 & 0 & 1 & 1 \\ \hline
\end{tabular}
\caption{Equipment Allowable Substitution Matrix}
}
\end{table}

Equipment types can be combined into composite types. For example, a common composite type is a combination of two shorts pulled by a single truck tractor. Other composite types are three shorts or a long combined with a short. In general, the equipment assigned to a load can be substituted by a larger type as long as the origin and destination facility of the load can accommodate the new type.

\section{Gini Model}
\label{B}

An alternative approach to the Minimax metric is to use a metric inspired by the Gini coefficient to promote equality across the entire distribution of scheduler burdens. This avoids focusing only on the single most-burdened scheduler and instead seeks to minimize the total pairwise inequality.

A hybrid model combining this Gini-based metric with the utilitarian objective of minimizing total changes can be formulated. Let the burden for each scheduler $s \in \mathcal{S}$, denoted $B_s$, be the total number of changes assigned to it:
\begin{equation}
B_s = \sum_{a \in \mathcal{A}_s} (1 - x_a \Phi_0(a)),
\end{equation}
where $\mathcal{A}_s$ is defined as in the Minimax model, representing the set of arcs under the responsibility of scheduler $s$, $x_a$ is the binary decision variable indicating whether arc $a$ is kept ($x_a = 1$) or removed ($x_a = 0$), and $\Phi_0(a)$ denotes the original arc assignment.

The Stage 2 model becomes:
\begingroup
\allowdisplaybreaks
\begin{align}
\label{eq:GiniStage2Objective}
\Delta^{fair} = \min \ & (1 - \omega) \sum_{s \in \mathcal{S}} B_s + \omega \sum_{\substack{s_1, s_2 \in \mathcal{S} \\ s_1 < s_2}} |B_{s_1} - B_{s_2}| \\
\text{s.t.} \quad
& \sum_{n \in \mathcal{N}} \sum_{r \in \mathcal{R}} I_{nr} \leq I^*, \label{eq:GiniConstraint1} \\
& \text{Constraints } (3.3) \text{ to } (3.6). \notag
\end{align}
\endgroup

where $\omega \in [0, 1]$ is a weight parameter that controls the trade-off between overall efficiency (minimizing total changes, when $\omega = 0$) and fairness (minimizing the sum of absolute differences, when $\omega = 1$). The term $\sum |B_{s_1} - B_{s_2}|$ is directly proportional to the Gini coefficient of the burden distribution.

To linearize the model, introduce continuous difference variables $D_{s_1 s_2}$ to represent the absolute differences between pairs of scheduler burdens:

\begingroup
\allowdisplaybreaks
\begin{align}
\allowdisplaybreaks
\Delta^{fair} = \min \ & (1 - \omega) \sum_{s \in \mathcal{S}} B_s + \omega \sum_{\substack{s_1, s_2 \in \mathcal{S} \\ s_1 < s_2}} D_{s_1 s_2} \label{eq:LinearGiniObjective} \\
\text{s.t.} \quad
& B_s = \sum_{a \in \mathcal{A}_s} (1 - x_a \Phi_0(a)), && \forall s \in \mathcal{S}, \label{eq:LinearGiniConstraint1} \\
& D_{s_1 s_2} \geq B_{s_1} - B_{s_2}, && \forall s_1 < s_2, \label{eq:LinearGiniConstraint2} \\
& D_{s_1 s_2} \geq B_{s_2} - B_{s_1}, && \forall s_1 < s_2, \label{eq:LinearGiniConstraint3} \\
& \text{Constraints } (\ref{Stage 1 Constraints 1}) \text{ to } (\ref{Stage 1 Constraints 4}), (\ref{eq:GiniConstraint1}). \notag
\end{align}
\endgroup

In addition to the minimax metric, tests are also conducted using the Gini index. Table \ref{tab:Stage 2 for the Gini Model} compares the three strategies: Efficiency (minimizing total changes), Fairness (minimax), and Gini (minimizing inequality in changes using the Gini index). The results reveal distinct trade-offs between the three models. For all classes, the Fairness and Gini approaches yield better results. The $Z$ metric significantly decreases compared to the Efficiency approach.  However, the $\Delta^{FAIR}$ significantly increases compared to the $\Delta^{GINI}$, which remains closer to $\Delta$.

\begin{table}[ht!]
  \centering
  \scalebox{0.81}{
  \caption{Stage 2 for the Gini Model}
  \label{tab:Stage 2 for the Gini Model}
    \begin{tabular}{c|c|c|c|c|c|c|c|c|c|c|c|c}
    \toprule
    \multirow{2}[4]{*}{\textbf{Class}} & \multirow{2}[4]{*}{\textbf{\# Sched.}} & \multirow{2}[4]{*}{\textbf{I$_0$}} & \multirow{2}[4]{*}{\textbf{I$_*$}} & \multicolumn{3}{c|}{\textbf{Efficiency}} & \multicolumn{3}{c|}{\textbf{Fairness}} & \multicolumn{3}{c}{\textbf{Fairness Gini}} \\
\cmidrule{5-13}               &       &       &       & \textbf{$\Delta^*$} & \textbf{$Z^*$} & \textbf{Time} & \textbf{$\Delta^{FAIR}$} & \textbf{$Z^{FAIR}$} & \textbf{Time} & \textbf{$\Delta^{GINI}$} & \textbf{$Z^{GINI}$} & \textbf{Time} \\
    \midrule
    1 & 2 & 4236 & 3444 & 321 & 199 & 1.44 & 357 & 188 & 1.53 & 336 & 204 & 1.40 \\
    \midrule
    2 & 5 & 10644 & 8357 & 952 & 354 & 4.74 & 1135 & 242 & 5.50 & 1085 & 286 & 8.42 \\
    \midrule
    3 & 8 & 18370 & 14709 & 1542 & 452 & 8.27 & 1852 & 274 & 11.30 & 1829 & 318 & 23.80 \\
    \midrule
    4 & 11 & 22733 & 17667 & 2048 & 443 & 12.25 & 2796 & 285 & 15.50 & 2563 & 330 & 49.5 \\
    \midrule
    5 & 15 & 28879 & 22863 & 2701 & 486 & 15.14 & 4184 & 308 & 18.16 & 3344 & 312 & 104.80 \\
    \midrule
    6 & 20 & 40799 & 32641 & 3725 & 710 & 22.13 & 7350 & 397 & 24.98 & 5149 & 400 & 88.71 \\
    \midrule
    7 & 30 & 58630 & 47863 & 4941 & 688 & 34.99 & 9039 & 331 & 39.65 & 7193 & 336 & 146.65 \\
    \midrule
    8 & 40 & 78488 & 65655 & 5943 & 732 & 44.33 & 13510 & 394 & 53.11 & 9237 & 397 & 209.02 \\
    \bottomrule
    \end{tabular}
}
\end{table}

Computation times highlight another key difference. While Efficiency and Fairness remain computationally reasonable for all classes, the Gini model can be more expensive. For all classes, enforcing fairness through the Gini model significantly increases the execution time compared to the efficiency and minimax approaches. For example, in Class 8, runtime rises from 53.11 (Efficiency) to 209.02 (Gini index), while increasing $Z$ from 394 to 397. This suggests that minimizing inequality is computationally challenging, especially for larger problems with more schedulers.

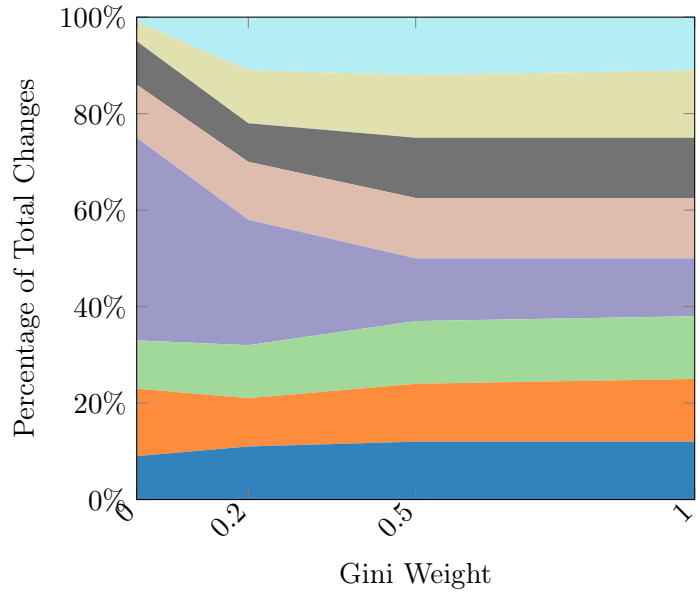
\begin{figure}[ht!]
\centering
\begin{tikzpicture}
\begin{axis}[
    ylabel={Percentage of Total Changes},
    xlabel={Gini Weight},
    width=9cm,
    height=8cm,
    xmin=0, xmax=1,
    ymin=0, ymax=100,
    xtick={0.0, 0.2, 0.5, 1.0},
    ytick={0, 20, 40, 60, 80, 100},
    yticklabel={\pgfmathprintnumber{\tick}\%},
    xticklabel style={rotate=45, anchor=east},
    stack plots=y,
    area style,
    enlarge x limits=false,
]
\addplot [fill=myblue, draw=none] coordinates {(0, 9) (0.2, 11) (0.5, 12) (1.0, 12)} \closedcycle;
\addplot [fill=myorange, draw=none] coordinates {(0, 14) (0.2, 10) (0.5, 12) (1.0, 13)} \closedcycle;
\addplot [fill=mylightgreen, draw=none] coordinates {(0, 10) (0.2, 11) (0.5, 13) (1.0, 13)} \closedcycle;
\addplot [fill=mypurple, draw=none] coordinates {(0, 42) (0.2, 26) (0.5, 13) (1.0, 12)} \closedcycle;
\addplot [fill=mytan, draw=none] coordinates {(0, 11) (0.2, 12) (0.5, 12.5) (1.0, 12.5)} \closedcycle;
\addplot [fill=mygray, draw=none] coordinates {(0, 9) (0.2, 8) (0.5, 12.5) (1.0, 12.5)} \closedcycle;
\addplot [fill=mykhaki, draw=none] coordinates {(0, 4) (0.2, 11) (0.5, 13) (1.0, 14)} \closedcycle;
\addplot [fill=mycyan, draw=none] coordinates {(0, 1) (0.2, 11) (0.5, 12) (1.0, 11)} \closedcycle;
\end{axis}
\end{tikzpicture}
\caption{Weighted Models Gini}
\label{fig:weighted models Gini}
\end{figure}

Figure \ref{fig:weighted models Gini} illustrates the effect of increasing the Gini weight ($\omega$) on the distribution of changes among eight schedulers, expressed as a percentage of total changes. The x-axis represents the Gini weight ranging from 0.2 (low emphasis on fairness) to 1.0 (high emphasis on fairness), while the y-axis shows the percentage share of changes for each scheduler. The goal of applying the Gini index is to reduce inequality by ensuring a more balanced distribution of changes among schedulers.

At $\omega=0.2$, the distribution is highly uneven, with Scheduler 4 dominating the share of total changes (close to 35–40\%), while some schedulers, like 1 and 2, have relatively small shares (around 10\% each). This indicates that when fairness is given little importance, a few schedulers absorb most disruptions, which minimizes overall cost but creates inequity. As $\omega$ increases to 0.5, the distribution becomes noticeably more balanced: Scheduler 4’s share drops significantly, while Schedulers 1, 2, and 3 gain a larger proportion, suggesting that fairness constraints start redistributing changes from heavily burdened schedulers to those previously less affected.

By $\omega=1.0$, the distribution appears relatively uniform, with all schedulers contributing between approximately 10\% and 15\% of total changes. This confirms that the Gini-based fairness mechanism successfully enforces equity among schedulers, preventing any single one from being disproportionately impacted. However, achieving this balanced state typically requires increasing the total number of changes, as seen in earlier tables, which means higher overall disruption. The trend demonstrates the trade-off managed by the Gini approach: it does not aim to minimize the maximum load (like minimax fairness) but reduces disparities across schedulers, leading to a more proportionate workload distribution at higher fairness weights.
\end{appendices}

\end{document}